\documentclass[conference]{IEEEtran}
\usepackage{times}

\usepackage[numbers]{natbib}
\usepackage{multicol}

\usepackage{amsmath} %
\usepackage{changes} %
\usepackage{amssymb}  %
\usepackage{amsfonts,bm}
\usepackage{siunitx}
\usepackage{xspace}
\usepackage[export]{adjustbox}
\usepackage{booktabs, array}
\usepackage{blindtext} 
\usepackage{soul}
\usepackage{lib} 
\usepackage{alias} 
\usepackage{multicol,multirow} 
\usepackage{algpseudocode}
\usepackage{algorithm}
\usepackage{cellspace}
\usepackage{subcaption,dcolumn}
\usepackage[normalem]{ulem}

\usepackage[breaklinks=true,letterpaper=true,colorlinks=true,citecolor=blue,bookmarks=true]{hyperref}

\usepackage[labelfont=bf]{caption}
\begin{document}

\title{
Opening Articulated Structures \\ in the Real World} %

\author{Arjun Gupta $\qquad$Michelle Zhang$^\star$ $\qquad$ Rishik Sathua$^\star$ $\qquad$Saurabh Gupta \\ University of Illinois at Urbana-Champaign \\ \url{https://arjung128.github.io/opening-articulated-structures}}

\makeatletter
\let\@oldmaketitle\@maketitle%
\renewcommand{\@maketitle}{\@oldmaketitle%
  \vspace{0.075in}
  \insertW{1.0}{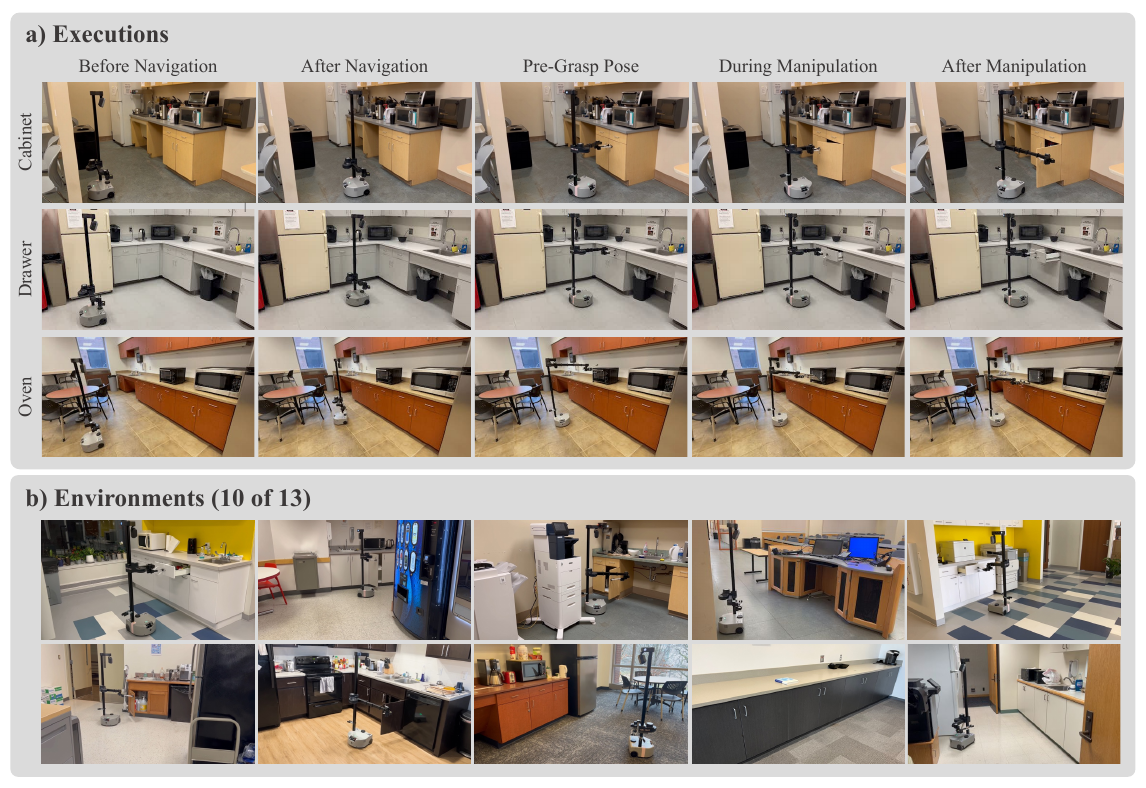}
  \centering
  \captionof{figure}{
  What does it take to build mobile manipulation systems with strong
  generalization capabilities, \ie the ability to competently operate on
  previously unseen objects in previously unseen environments? 
  This work seeks to answer this question using \textit{end-to-end opening 
  of articulated structures without any privileged information} as a mobile manipulation testbed.
  Specifically, we develop \system, a MOdular System for
  opening ARTiculated structures, and conduct extensive testing of
  the end-to-end system in real environments. \textbf{a)} shows executions of
  \system opening novel cabinets, drawers, and ovens in unseen environments.
  \textbf{b)} shows the other 10 unseen environments (across 10 buildings) used in
  our large-scale real world study. Our findings provide insights to
  researchers and practitioners aiming to build generalizable mobile
  manipulation systems.}
  \figlabel{teaser}}
  \vspace{-0.3in}
  \bigskip 
\makeatother
\maketitle
\addtocounter{figure}{-1}

\thispagestyle{empty} %
\pagestyle{empty} %

\begin{abstract}

What does it take to build mobile manipulation systems that can competently
operate on previously unseen objects in previously unseen environments? This work answers
this question using \textit{opening of articulated structures} as a 
mobile manipulation testbed. 
Specifically, our focus is on the \textit{end-to-end} performance on this task \textit{without any privileged information}, \ie the robot starts at a location with the novel target articulated object in view, and has to approach the object and successfully open it.
We first develop a system for this task, and then conduct 100+ end-to-end system tests
across 13 real world test sites. Our large-scale study reveals a number of surprising findings: a) modular systems outperform end-to-end learned systems for this task, even when the end-to-end learned systems are trained on 1000+ demonstrations, b) perception, and not precise end-effector control, is the primary bottleneck to task success, and c) state-of-the-art articulation parameter estimation models developed in isolation struggle when faced with robot-centric viewpoints. Overall, our findings highlight the limitations of developing components of the pipeline in isolation and underscore the need for system-level research, providing a pragmatic roadmap for building generalizable mobile manipulation systems.
Videos, code, and models are available on the project website: \url{https://arjung128.github.io/opening-articulated-structures/}.

\end{abstract}

\IEEEpeerreviewmaketitle

\section{Introduction}
\seclabel{intro}

Developing mobile manipulators that can reliably perform everyday tasks in diverse environments remains a fundamental challenge in robotics.
A major obstacle to realizing this vision lies in the lack of strong generalization capabilities: current systems struggle to adapt to novel objects and unfamiliar settings.
This difficulty stems from the inherent complexity of mobile manipulation in arbitrary environments, which requires robust perception, precise motion planning, and successful execution.
While considerable efforts have been devoted to improving each of these subproblems in isolation, relatively few studies bring these components together into a cohesive system designed for real world deployment.
Building and studying end-to-end systems rather than isolated subproblems can reveal hidden challenges crucial for practical deployment. 
By focusing on the full system, research can shift attention towards overall task success rather than marginal gains in individual components that may not be the primary bottleneck.
Therefore, prioritizing system-level end-to-end performance—particularly the ability to generalize to unseen objects in unfamiliar settings—is essential for achieving reliable and capable mobile manipulators in everyday environments.

To tackle the challenge of strong generalization within the context of mobile manipulation, we work on the task of opening articulated structures in diverse previously unseen real world environments (\figref{teaser}).
Articulated structures, such as cabinets, drawers, and ovens, are ubiquitous in indoor environments, making their reliable manipulation a key capability for robots operating in the wild. 
In our problem setting, a Stretch RE2 robot (a commodity mobile manipulator without any hardware modifications) is placed in a novel environment in front of a previously unseen articulated object, and the objective is to solve the full end-to-end task. This includes detecting the object and estimating its articulation parameters, generating a precise whole-body motion plan, navigating to an optimal interaction position, securing a firm grasp on the handle, and successfully opening the object without collisions—all performed \textit{zero-shot, without any privileged information about the object or environment}.
Thus, this problem setting includes all of the key challenges that a mobile
manipulator would face when interacting with previously unseen objects in
unfamiliar environments in the real world, and thus forms a good test bed for
study. We first developed an end-to-end mobile manipulation system to tackle
this task and then conducted extensive end-to-end testing to understand the
current bottlenecks in building such a system. This paper describes the system
we build, the testing we conducted, and the lessons we learned along the way.

\noindent {\bf The System.} We considered two broad ways of putting together
such a system: a modular approach and an end-to-end learning approach, but
ultimately favored a modular approach. Our approach, called \system for a
MOdular System for opening ARTiculated structures, develops and utilizes
state-of-the-art modules for perception, planning, and adaptation (shown in
\figref{pipeline}). Very briefly, \system adapts a Mask RCNN
model~\cite{he2017mask} for inferring articulation parameters, extends a
trajectory optimization framework for producing whole body motion
plans~\cite{gupta2023predicting}, and utilizes proprioceptive feedback for
mitigating last centimeter errors.

\noindent {\bf The Testing.}
We conduct large-scale tests to assess the capabilities of \system.  This
testing is conducted  in 13 test sites from 10 different buildings across 31
different articulated objects in the wild.  Testing sites include offices,
classrooms, apartments, office kitchenettes, and lounges (see~\figref{teaser}).
None of these test objects or sites were used for development in any way, and
{\it testing was done just once} to mitigate any influence on the design of the
system. Each trial started with the robot being placed such that the object was
in view. A user selected which articulated object was to be opened and hit the
go button. If the drawer was opened more than 24cm or a cabinet was pulled open
by {$60^\circ$}, the trial was deemed a success.
Finally, we also conduct experiments to understand a) how \system compares
to an end-to-end learning approach, b) how sensitive \system is to the 
performance of each individual submodule, c) whether \system can generalize
to diverse handles and other articulation types, and d) what are the biggest
bottlenecks that cause \system to fail, 
providing insights into open challenges in building generalizable mobile manipulation systems.

\begin{figure*}[tbp]
\insertW{1.0}{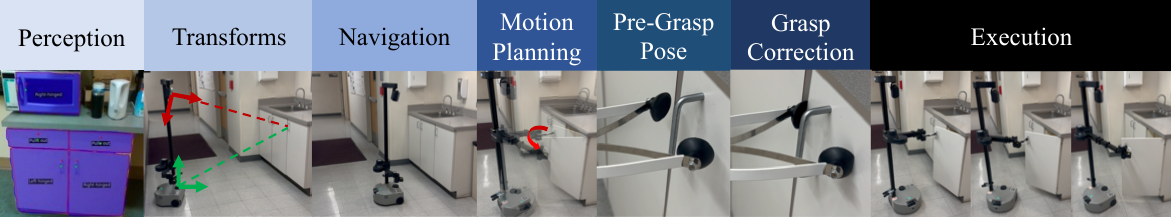}
\caption{\textbf{\system Design.} 
The perception module outputs 
3D articulation parameters in the robot frame using RGB-D images.
The robot then navigates to the target location based on articulation type.
Next, we use SeqIK to find a whole-body motion plan. We execute the first robot configuration from the plan to obtain a pre-grasp pose. We then use our contact-based adaptation for improved grasping. Once the handle is grasped, we execute the rest of the plan.}
\figlabel{pipeline}
\end{figure*}

\noindent {\bf The Findings.} Our large-scale system-level study revealed various takeaways:

\noindent \textbullet~\textbf{A modular system outperforms an end-to-end learning system when tested
for generalization to unseen objects in novel environments.} 
We compare \system to the recent and contemporary imitation
learning work Robot Utility Models (RUM) from Etukuru \etal~\cite{etukuru2024robot}. RUM is
trained on a large-scale dataset of expert demonstrations: 1200 demos for
opening cabinets, and 525 demonstrations for opening drawers across
approximately 40 environments, by far the largest imitation learning dataset
for this task.  To our surprise, we find that the modular system outperforms
this latest end-to-end learning method (\secref{comparisons} and
\tableref{rum}). 
This result is particularly useful in the context of the ongoing debate between
modular learning and end-to-end learning in robotics at large, and specifically
for recent end-to-end learning works that tackle mobile manipulation
problems~\cite{shafiullah2023bringing, xiong2024adaptive, yang2023harmonic,
gu2023maniskill2}.\\
\textbullet~\textbf{State-of-the-art perception modules for articulation parameter
prediction struggle on images from a robot.}
Significant progress has been made in predicting articulation parameters from
images \cite{sun2023opdmulti, opd, chen2024urdformer, qian2023understanding,
Mo_2021_ICCV, wu2022vatmart, ning2023learning, morlans2023aograsp}.
However, most of this progress has been made in isolation, without evaluating
these methods as part of a complete, end-to-end system on real-world,
in-the-wild images.  When tested on viewpoints from the robot in the real
world, recent state-of-the-art approaches such as
OPD-Multi~\cite{sun2023opdmulti} and AO-Grasp~\cite{morlans2023aograsp}
struggle to make accurate predictions.  In contrast, our choice to
develop an Articulation-parameter Prediction Module (APM) targeted towards images
that the robot was expected to see, achieves substantially better performance \vs
OPD-Multi~\cite{sun2023opdmulti} (\secref{apm-eval} and \figref{opdmulti}) and
AO-Grasp~\cite{morlans2023aograsp} (\secref{apm-eval},
\tableref{ao-grasp-eval}, \figref{ao-grasp}).\\
\textbullet~\textbf{Modular systems can generalize to diverse handles and be easily
extended to other articulation types.}
Not only did the modular system \system outperform the competing
imitation learning method RUM~\cite{etukuru2024robot}, we found the modular
system to be advantageous in other ways.
\system is able to generalize to diverse handles (\secref{handles}).
As vision foundation models keep improving, a modular system continues to
automatically improve.  Additionally, we found it easy to extend our modular
system to a new articulation type (horizontal-hinged toaster ovens,
\secref{ovens}). In comparison, an imitation learning system will need to
recollect a large amount of training data for tackling a new articulation type. \\
\textbullet~\textbf{The failure mode analysis reveals that robust perception, and not
control, is the biggest bottleneck in overall task success.} Surprisingly,
perception and not the precise
control of the end-effector (to follow the handle as the object articulates) is
what makes this problem hard, contrary to what prior work focuses
on~\cite{gupta2023predicting,karayiannidis2016adaptive,chitta2010planning}.
Inability to detect objects and handles accounts for 59\% of the failures of the
system (\figref{failures}, \secref{failure_mode_analysis}). Furthermore, the 
specific ways in which the perception system failed
was also revealing. It is not as much a failure in estimating articulation
parameters, but the detection of target objects and estimation of the handle
location in 3D are the bottlenecks. Specifically, cabinets with meshed surfaces
do not get detected and keyholes get mistaken as handles. It may be important to
capture data for such corner cases.\\ 
\textbullet~\textbf{Last-centimeter grasping remains challenging.} 
Control was 
surprisingly robust to mis-estimations in the articulation parameters. Once the
end-effector acquired a firm grasp of the handle, we found that the system
is able to sufficiently open the cabinet even when the radius is off by as much
as $10cm$ (\secref{robustness}).
However, securing this initial grasp proved to be somewhat challenging, with last-centimeter grasping errors accounting for 25\% of all failures.
While proprioceptive feedback provided a simple yet effective adaptation mechanism, it did not fully eliminate last-centimeter grasping errors.
Tackling such last-centimeter errors remains an avenue for future work.
Some form of a closed-loop visual grasping controller that is robust to slight mis-calibration and imprecise navigation may improve performance on mobile manipulation tasks which require precise grasping.

\noindent \ul{\textbf{Broader Implications.}}
Our findings provide key insights for designing generalizable mobile manipulation systems, offering several takeaways that should inform future research and system development.
In particular, our large-scale, real-world experiments show that vision models trained in isolation frequently underperform when confronted with the distinct viewpoints encountered by robots, highlighting the need for comprehensive, system-level research.
Moreover, we identify perception as the principal bottleneck in achieving consistent task success, a finding that underscores the need for future research to prioritize robust perception modules.
Additionally, our results indicate that modular approaches tend to outperform end-to-end methods, cautioning practitioners that scaling imitation-learning datasets to even 1,000 demonstrations does not necessarily lead to broad generalization.
Notably, modular systems also facilitate easier adaptation to related tasks without requiring extensive new data collection.
Finally, last-centimeter grasping remains a key challenge. 
While contact correction helps alleviate this failure mode to some extent, developing closed-loop interaction strategies could improve performance in tasks requiring precision, making it an important direction for future research.
Overall, these lessons provide a pragmatic roadmap for researchers and practitioners aiming to build generalizable mobile manipulation systems.

\begin{figure*}[t]
\insertW{1.0}{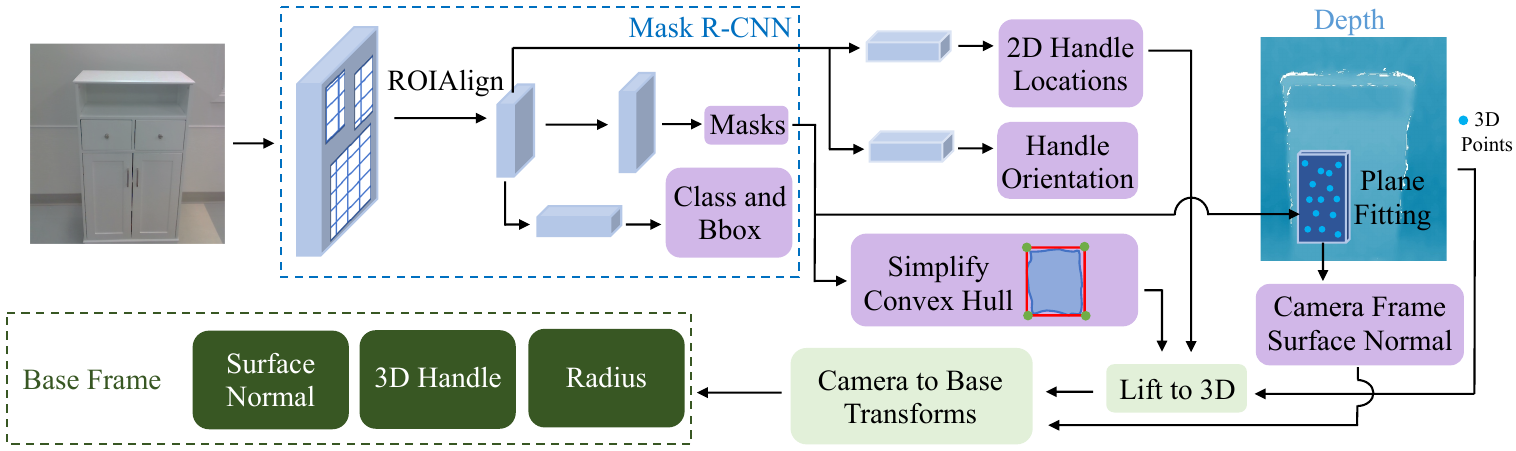}
\caption{\textbf{Overview of the Articulation-parameter Prediction Module (\model).} Given an RGB image our modified Mask RCNN detects articulated objects and predicts the articulation type, the handle orientation, the 2D segmentation mask, and the 2D handle keypoint. We fit a convex hull to the segmentation mask and simplify it to a quadrilateral. We fit a plane to the depth image points that lie inside the mask to estimate the surface normal. The 2D handle and quadrilateral corners are lifted to 3D using the depth image. All predictions are transformed to the robot base frame. 
The final output of the module includes the 3D handle coordinate and surface normal in the base coordinate frame for all articulated objects, and additionally the radius and rotation axis for cabinets.} 
\figlabel{perception}
\end{figure*}

\section{Related Work}
\seclabel{related}

\subsection{Predicting Articulation Parameters}
Researchers have extensively looked at different aspects: a) construction of various datasets (from
simulation~\cite{ai2thor, 
ehsani2021manipulathor,
gan2020threedworld}, real world images \cite{sun2023opdmulti, opd, chen2024urdformer}, and real
world 3D scans~\cite{gupta2023predicting, torne2024rialto}), b) use of different input
modalities to make predictions (\rgb images~\cite{sun2023opdmulti}, point
clouds~\cite{Liu_2023_ICCV, nie2022sfa, wang2024articulated, EisnerZhang2022FLOW, zhang2023flowbotplus, li2019articulated-pose, schiavi2023learningagentawareaffordancesclosedloop}, \rgbd images~\cite{opd, xu2022umpnet, pmlr-v100-abbatematteo20a, zeng2021visualidentificationarticulatedobject}), c) use of diverse
sources of supervision~\cite{qian2023understanding}, and d) predicting sites
for interaction (\ie handles) in addition to articulation
parameters~\cite{Mo_2021_ICCV, wu2022vatmart, ning2023learning, yu2024gamma}. 
Most past works are largely evaluated in simulation, and because no prediction
is made for other articulation parameters (\eg radius), these works cannot directly be
used to generate motion plans to fully articulate objects in the real world 
without any privileged information.
Sun \etal~\cite{sun2023opdmulti} augment the Mask RCNN architecture by adding 
additional heads specifically designed to predict articulation parameters in 3D.
We also add additional heads to Mask RCNN; however, rather than directly 
predicting 3D outputs from the \rgbd input, we adopt a two-stage approach 
involving {\it 2D prediction from \rgb images} followed by {\it 3D lifting using the depth image}~\cite{gupta2015aligning,xie2023part}.

\subsection{Generating Motion Plans}
Opening doors and drawers requires the
end-effector to conform to the constraint defined by the object handle's
trajectory (which in turn is determined by the location of the articulation
joint and the handle). Depending on the robot morphology, robot configurations
that conform to this end-effector constraint might represent a measure zero
set. This makes it challenging to directly use sampling-based motion
planners~\cite{kavraki1996probabilistic, kuffner2000rrt}. Past research has
therefore developed specialized methods for planning under such
constraints~\cite{kingston2018sampling, Berenson2018} and used it to articulate
objects~\cite{berenson2011task, chitta2010planning, ruhr2012openingdoors,
peterson2000high, meeussen2010autonomous, Burget2013wbmparticulated,
Vahrenkamp2013irm, honerkamp2021learning, honerkamp2023learning, narayanan2015task}. Another line of work casts it as a trajectory
optimization~\cite{gupta2023predicting, zucker2013chomp, schulman2014motion} or
optimal control ~\cite{Farshidian2017brealtimeplan, Pankert2020perceptivempc,
sleiman2021unified, mittal2021articulated} problem.  As these only search for a
solution locally (\vs motion planning that searches globally via sampling), it
is important to properly initialize the trajectory optimizers.  Recent work
from Gupta \etal~\cite{gupta2023predicting} alleviates this limitation by using
learning to predict good initializations for trajectory optimization thus
generating high-quality motion plans quickly. We adopt their approach but
extend it to produce \textit{whole body motion plans} as we describe in
\secref{approach}.

\subsection{Mobile Manipulation}
Recent papers have looked at different aspects:
pick-move-place tasks~\cite{yenamandra2023homerobot}, high-level planning given
natural language instructions~\cite{saycan2022arxiv}, dynamic whole body
control~\cite{fu2022deep}, building simulators~\cite{robocasa2024}, 
developing tele-operation setups~\cite{fu2024mobile}. 
\cite{tri_mm_paper} tackle a grocery shopping scenario with a custom
robot, and conduct extensive field tests over 18 months. 
Many recent works adopt an end-to-end learning approach for opening articulated
objects \cite{yang2023harmonic, shafiullah2023bringing, xiong2024adaptive, ito2022efficient,
etukuru2024robot}.
\cite{yang2023harmonic} also work with a Stretch Robot but focus on 
sim2real transfer and modify the
environment to simplify the challenges due to closed-kinematic chains.
The Dobb-E
system from \cite{shafiullah2023bringing} showcases interactions with
articulated objects, but requires retraining on each test object using
human-collected demonstrations.
Work from \cite{xiong2024adaptive} replaces the need for test-time
demonstrations with test-time adaptation via RL requiring about an hour
of interaction for the adaptation. \cite{Sleiman_2023} enable a quadruped
to articulate heavy doors and dishwashers, but assume privileged
environment information (\eg a model for the door).
Recently, Etukuru \etal~\cite{etukuru2024robot} introduced RUM, a
large-scale imitation learning system for opening articulated objects,
collecting over 1,000 demonstrations across approximately 40 environments.
However, their approach relies on ground-truth articulation type, an
eye-in-hand camera, an approximate handle height, and requires the robot to be
positioned directly in front of the object with a clear view of the handle.
In contrast to these approaches, we develop a system that operates on 
novel object instances in novel environments in a {\it zero-shot} manner
{\it without requiring any privileged information}.

\subsection{Modular Learning}

Methods that integrate learning-based components with classical methods, have emerged as an effective approach for building robust robotic systems.
Examples span a wide variety of domains: grasping \cite{mousavian2019graspnet, dexnet2, cartman, lu2020multifingered}, aerial robotics \cite{scaramuzza2014vision}, and autonomous driving \cite{pmlr-v87-mueller18a, ijcai2017p661}.
For instance, DexNet 2.0 \cite{dexnet2} uses a neural network to evaluate grasp quality and a sampling-based planner to execute the chosen grasp.
A large body of work in robotic navigation also utilizes modular learning \cite{chaplot2020learning, chaplot2020object, ramakrishnan2020occant, chaplot2020neural, ramakrishnan2022poni, chang2020semantic, gervet2023navigating}.
Notably, sharing similarities with our work, Gervet \etal~\cite{gervet2023navigating} design a modular system for navigation, and via a large real world evaluation, demonstrate the superior performance of such a system relative to classical approaches and end-to-end learning.
Recently, modular learning approaches have also leveraged vision-language models (VLMs) \cite{huang2024rekep} and have been applied successfully to humanoid control \cite{he2024omnih2o, lin2025sim}.
For example, ReKep \cite{huang2024rekep} employs vision foundation models for high-level reasoning, followed by trajectory optimization for low-level control.
While modular learning methods vary in the choice of modules and their reliance on learning, in this work, we present a practical and effective modular design demonstrated on the challenging mobile manipulation task of opening articulated objects in-the-wild.

\begin{figure}[t]
\insertW{0.5}{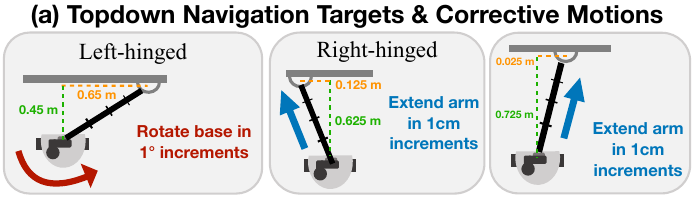}
\caption{{\bf Topdown Navigation Targets and Corrective Motions.} 
We show the topdown navigation targets relative to the handle for each articulation type.
For left-hinged cabinets, correction is a rotation in $1^\circ$ increments. For the other objects, we extend the arm in $1cm$ increments.}
\figlabel{corrective_vertical}
\end{figure}

\section{Modular System for Opening Articulated Objects (\system)}
\seclabel{approach}

We follow a modular approach comprising of a) a Perception Module that predicts
articulation parameters using an on-board RGB-D camera, b) a Motion Plan
Generator that converts predicted articulations into whole-body motion plans,
and c) an Execution Module that adapts and executes the generated motion plan
using proprioceptive feedback. \figref{pipeline} shows an overview.
We describe the pipeline in terms of two prototypical articulated
objects: drawers for prismatic joints and cabinets for hinged joints, but our
pipeline is more general as our experiments will reveal.

\subsection{Predicting articulation parameters using on-board \rgbd images (\model)}
\seclabel{maskrcnn}
Our Articulation-parameter Prediction Module (\model), shown in \figref{perception}, detects and predicts
articulation parameters for cabinets and drawers from RGB-D images.
These articulation parameters include the 3D handle location and surface normal
for drawers, and additionally the rotation axis and radius for cabinets. 
We predict 2D quantities from RGB images, and lift them to 3D 
using the depth image.

For 2D prediction from RGB images, we adopt Mask RCNN \cite{he2017mask}.  As
is, Mask RCNN predicts a 2D segmentation mask and the class of each detected
object (in our case, the articulation type: drawer, left-hinged cabinet, or
right-hinged cabinet). We add additional heads to Mask RCNN to predict the handle's 2D
pixel location and orientation (horizontal or vertical).  Both
additional heads are treated as classification tasks and trained with
a cross-entropy loss. For the 2D handle coordinate prediction, we minimize the
cross-entropy loss over a 2D spatial map.

We use the depth image to lift these 2D predictions to 3D.
For the surface normal, we fit a plane to the 3D points within 
the predicted segmentation mask. 
For the 3D handle position, we lookup the 3D coordinate corresponding to the 
predicted 2D handle coordinate.
For cabinets, we also need the radius and the axis of rotation. We compute the
convex hull of the predicted 2D segmentation mask, and simplify it to a
quadrilateral.
We lift the vertices of this quadrilateral to 3D and infer the rotation axis
from the corners, \eg for a left-hinged cabinet we define the rotation axis
using the left corners. We use the distance of the handle to its projection on
the rotation axis as the radius.

We train our modified Mask RCNN on the ArtObjSim dataset~\cite{gupta2023predicting}. 
ArtObjSim contains 3500+ articulated objects
across 97 scenes from the HM3D dataset \cite{ramakrishnan2021hm3d}. 
Each articulated object comes with 3D annotations for its extent, 
handle location, articulation type, and articulation parameters.
We train on images and 2D annotations rendered out from arbitrary 
locations in the scene.

\subsection{Motion plan generation}
Given waypoints computed using our predicted articulation parameters, our motion plan generator synthesizes a navigation target and a whole-body
motion plan to open the given articulated object in a collision-free manner.
These waypoints, representing end-effector poses, are determined based on the predicted articulation type and handle location: a linear trajectory along the predicted surface normal for drawers, and a quarter-circle path also using the predicted radius for cabinets.
We build upon past work~\cite{gupta2023predicting} that
converts ground-truth end-effector pose trajectories into robot joint angle
trajectories.
Specifically, rather than casting it as a constrained motion planning problem,
Gupta \etal~\cite{gupta2023predicting} view it as a trajectory optimization problem and design SeqIK, a
trajectory optimizer specifically suited to this task. SeqIK translates an 
initial robot configuration (base position and arm joint angles denoted by
$\theta_0$) into a \textit{strategy} that can be decoded into a motion plan
(desired joint angle trajectory) when provided with a
desired end-effector pose trajectory $\bfw$, via 
$\text{SeqIK}(\theta_0)(\bfw)$. 
SeqIK performs inverse kinematics calls sequentially, warm-starting the next
inverse kinematics call with the output of the current.
This leads to accurate motion plans with only a few IK calls.

\begin{figure*}[t]
\insertW{1.0}{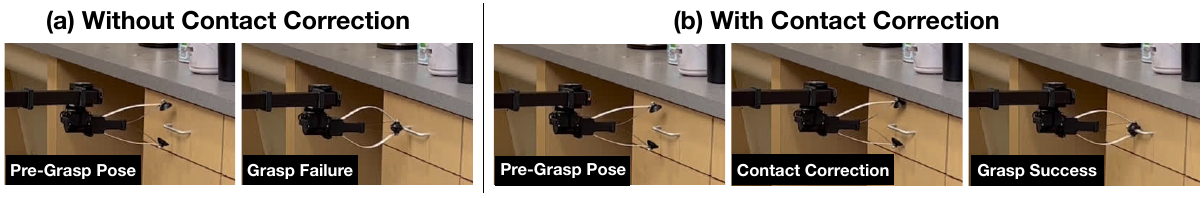}
\caption{{\bf Contact Correction.} (a) shows a grasping attempt with \textit{No contact correction}, whereas (b) shows the grasping attempt  \textit{with contact correction}. Due to compounding errors and shrinkage of the gripper when closing, the {\it without contact correction} version fails to grasp, whereas our contact-based correction mechanism leads to a successful grasp.}
\figlabel{contact}
\end{figure*}

We extend the framework from~\cite{gupta2023predicting} in
three ways. First, \cite{gupta2023predicting} works with the 
Franka Emika 
Panda robot. We adopt their implementation to work with the 
Stretch RE2 robot, which has fewer degrees of freedom.
Second, we work with articulation parameters predicted using 
\model as opposed to ground truth articulation parameters. 
Finally, we obtain \textit{whole-body} motion plans rather than base placement +
arm-only plans developed in \cite{gupta2023predicting}. We find whole-body
planning essential to fully open a wide variety of cabinets and drawers due to
the limited number of degrees of freedom of the Stretch RE2.  Here, in each
SeqIK step, given the fixed initial base pose, we allow inverse kinematics
to search for base rotation in addition to the arm joints.

SeqIK requires an initial base position and arm joint angles $\theta_0$. For
the initial base position, we utilize MPAO (No neural network), the data-driven
method from \cite{gupta2023predicting} that ranks robot configurations (base
positions and arm joint angles) by how successfully SeqIK can decode these
configurations into collision-free constraint-abiding motion plans.
\figref{corrective_vertical}(a)
shows the base positions found by this procedure for drawers,
left-hinged and right-hinged cabinets. We use these as the navigation targets
for each articulation type respectively. For the initial arm joint angles
$\theta_0$, we find that using the default neutral joint angles works well
for the Stretch RE2.

\subsection{Adapting and executing motion plans using proprioceptive feedback}
Minor errors in state estimation, inaccuracies in navigation, and imperfections 
in calibration can compound over time to ultimately prevent successful handle grasping.
Specifically, the system exhibits very little tolerance to errors along the depth direction.
This is further exacerbated by the hardware
design of the gripper that becomes shorter upon closure.
To combat this compounded error, we employ a \textit{contact-based correction strategy} aimed at refining the pre-grasp pose: specifically, we extend the gripper incrementally towards the object surface until physical contact is detected.
See \figref{corrective_vertical} for a 
visualization of the strategy for the different articulated objects. 
As is, the end-effector approaches the handles from the side, which 
works well for horizontal handles, but poses issues for vertical handles.
For objects with vertical handles, we thus have an additional vertical
primitive
before the 
contact-based correction mechanism. 
We use the arm effort signal and the gripper's yaw effort signal to detect
contact. The change in the end-effector pose during contact correction phase
provides feedback to update the desired trajectory and motion plan.
We ablate the use of contact correction in
\secref{adaptation-ablation}.

\begin{table}
  \centering
  \resizebox{\linewidth}{!}{
  \begin{tabular}{lrrrr}
  \toprule
                              & Drawer & Left-Hinged & Right-Hinged & Total\\
  \midrule
  RUM~\cite{etukuru2024robot}  & 4/6    & 1/7         & 1/7         & 6/20 \\
  \system (Ours) & 5/6    & 3/7         & 5/7     &  13/20 \\
  \bottomrule
  \end{tabular}}
  \caption{Comparison of \system \vs RUM~\cite{etukuru2024robot}, a recent large-scale end-to-end imitation learning method trained on 1200 demos for opening cabinets and 525 demos for opening drawers across 40 different environments. Our evaluation spans objects in-the-wild from four previously unseen environments across three buildings.}
  \tablelabel{rum}
\end{table}

\subsection{Full end-to-end execution}
Starting from where the robot can see the target articulated object,
we use \model to predict several key properties: the handle location (in the camera's coordinate frame), the surface normal of the surface, and the articulation radius (for cabinets).
We use a calibrated robot URDF to transform these 3D predictions
from the camera frame to the base frame.
Next, we generate a whole-body motion plan and execute the first qpose (full robot configuration), followed by our contact-based correction mechanism.
Once the handle has been grasped, the robot proceeds to execute the remainder of the precomputed motion plan to complete the task.

\renewcommand{\arraystretch}{1.2}
\begin{table*}
\centering
\caption{Accuracy of \model predictions on images collected during our large-scale real world evaluation.}
\tablelabel{maskrcnn_table}
\begin{tabular}{lcccccccccc}
\toprule
                            
                            & & \multicolumn{4}{c}{\bf \model with Mask RCNN} & &
                            \multicolumn{4}{c}{\bf \model with Detic} \\
                            \cmidrule(lr){3-6} \cmidrule(l){8-11}
                            & & \bf Drawer & \bf Left-Hinged & \bf Right-Hinged
                            & \bf All & & \bf Drawer & \bf Left-Hinged & \bf Right-Hinged & \bf All \\
\midrule
Detection                     & & 9/9 & 8/9 & 12/13 & 29/31 & & 9/9 & 9/9 & 13/13 & 31/31 \\
Handle orientation            & & 9/9 & 8/8 & 11/12 & 28/29 & & 9/9 & 9/9 & 12/13 & 30/31 \\
Articulation type             & & 9/9 & 8/8 & 12/12 & 29/29 & & 9/9 & 9/9 & 12/13 & 30/31 \\
Mean handle error             & & 3.17 cm & 1.43 cm & 1.35 cm & 2.11 cm & & 1.51 cm & 0.82 cm & 1.34 cm & 1.27 cm \\
Mean radius error             & & n/a & 0.47 cm & 1.32 cm & 0.98 cm & & n/a & 3.28 cm & 2.31 cm & 2.70 cm \\
\bottomrule
\end{tabular}
\end{table*}

\section{Experiments}
\seclabel{experiments}

We work with the Stretch RE2 robot.
We first present our end-to-end system test results, 
evaluating
\system on 31 novel drawers and cupboards across 10 buildings (\secref{end_to_end}).
To see how a modular system compares to an end-to-end learning approach, we compare 
\system to RUM \cite{etukuru2024robot} and a sim2real imitation learning approach (\secref{comparisons}).
To further understand the system, we then evaluate individual pipeline modules (\secref{individual_module_eval}).  
This includes evaluating the quality of our MaskRCNN-based perception
module (as well as a Detic-based perception model) on real world images, 
comparing \model to two recent articulation parameter prediction systems \cite{morlans2023aograsp, sun2023opdmulti},
and an ablation for the contact-based correction
mechanism. %
To understand the interaction between
perception and control, we quantify the robustness of execution to state
estimation inaccuracies (\secref{robustness}).
We then 
study the generalization of our
pipeline to other articulation types and diverse handles (\secref{ovens_and_handles}),
before we analyze the failure modes of our system (\secref{failure_mode_analysis}).

\subsection{End-to-end System Tests}
\seclabel{end_to_end}

We test our end-to-end system across 8 office buildings and 2 apartments on a
total of 31 distinct articulated objects. These test objects do not overlap
with ones used for development. 
In each test, the robot is placed approximately $1.5m$ from the target object
with the camera oriented so as to have the target object in view. We introduce
variation in the starting pose of the robot to test the robustness of the
approach but use the same starting pose when comparing different methods.  A
trial is successful if the drawer is pulled out by $24cm$ / the
cupboard is opened more than $60^\circ$.

\noindent \textbf{Results.} Overall, our system achieves a 61\% success rate
across 31 unseen cabinets and drawers in unseen real world environments.
Figure 1 shows examples of deployments of our full pipeline. For
most successful trials, the robot opens the drawer / cupboard completely (\ie
drawers by $35 cm$ and cupboards by $90^\circ$) in a graceful manner (see videos in supplementary materials).
\secref{failure_mode_analysis}
provides a extensive discussion of the failure modes.

\subsection{Comparisons to End-to-End Imitation Learning}
\seclabel{comparisons}

\subsubsection{\ul{Robot Utility Models (RUM)}}
Next, we investigate how \system compares to a large-scale end-to-end imitation
learning approach in generalizing to novel objects in previously unseen
environments. Specifically, we evaluate \system against Robot Utility Models
(RUM), recently introduced by Etukuru \etal~\cite{etukuru2024robot}.  RUM is
trained on an extensive dataset of expert demonstrations: 1,200 for opening
cabinets and 525 for opening drawers, collected across approximately 40
environments. This dataset is by far the largest imitation learning dataset for
articulated object manipulation, making RUM a strong baseline for our
comparison.

Unlike \system, RUM operates under several assumptions. It requires prior
knowledge of the articulation type (\eg drawer vs. cabinet), an approximate
handle height, and it assumes the robot is positioned directly in front of the
object with an optimal view of the handle. To overcome these assumptions, we
leverage \system's \model to infer the articulation type and handle height, and
we use \system's navigation module to position the robot directly in front of
the object (centered on the handle), among other minor adjustments described in \secref{rum_details}. 

As shown in \tableref{rum}, \system substantially outperforms RUM in a paired
evaluation on unseen objects across four novel environments.  
In addition to achieving higher success rates, \system is also faster than RUM (see \secref{rum_timing}).
One common failure
mode we observe for RUM is the inability to grasp the handle of the object,
despite getting very close to it.
Another is incorrect handle selection, particularly when multiple handles are
in close proximity and within view. 
This stems from RUM's inability to explicitly specify the task, 
making it susceptible to such ambiguities and highlighting a key limitation of the approach.
Furthermore, our evaluation focuses solely on in-the-wild objects, where such challenging cases 
naturally arise, whereas a considerable portion of RUM’s evaluation was conducted on isolated 
objects in controlled lab environments.
These findings highlight a) the
importance of end-to-end system tests in-the-wild, b) the challenging nature of this
problem that requires generalization to previously unseen objects in novel
environments, and c) the effectiveness of \system.

\begin{table}
  \centering
  \resizebox{\linewidth}{!}{
  \begin{tabular}{lrrrr}
  \toprule
                              & Drawer & Left-Hinged & Right-Hinged & Total\\
  \midrule
  AO-Grasp \cite{morlans2023aograsp}  & 13.11 cm    & 33.66 cm        & 24.46 cm     &  23.36 cm \\
  Ours & 3.17 cm    & 1.43 cm         & 1.35 cm         & 2.11 cm \\
  \bottomrule
  \end{tabular}}
  \caption{\textbf{Quantitative Comparison to AO-Grasp.} Mean handle error across all instances of our real world testing.}
  \tablelabel{ao-grasp-eval}
\end{table}

\begin{figure*}[t]
\insertW{1.0}{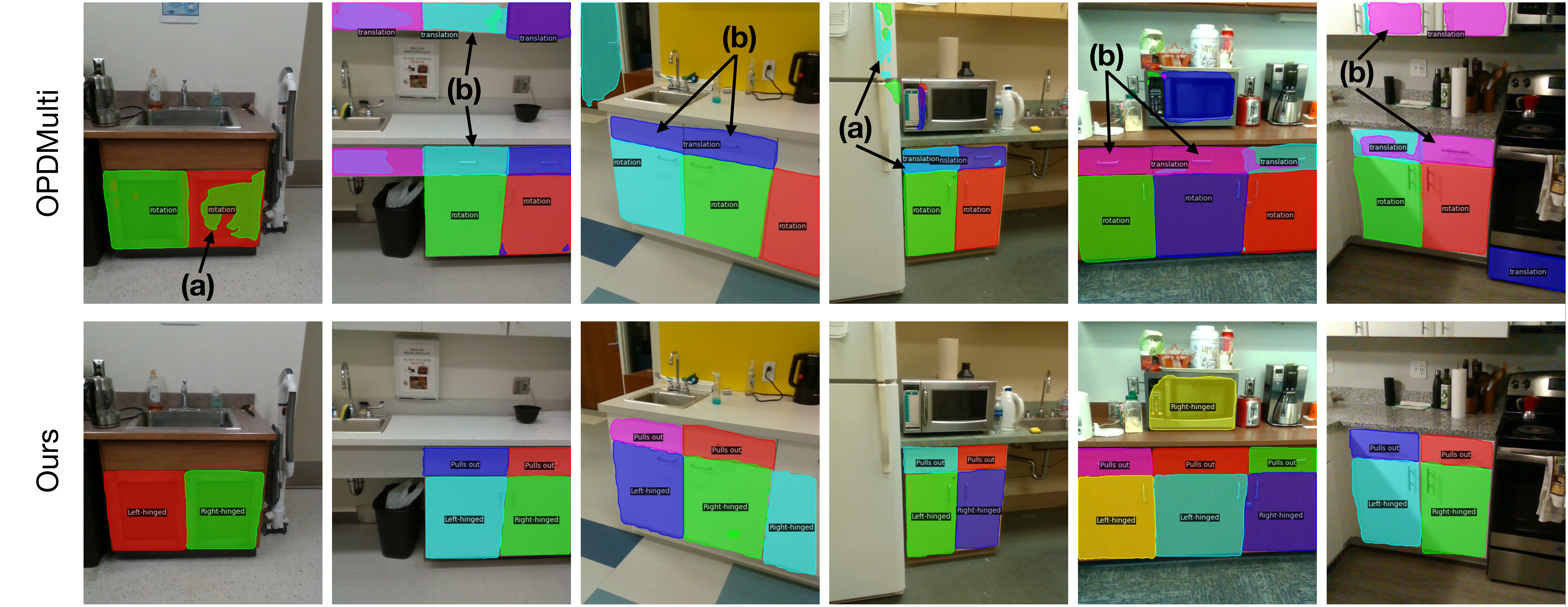}
\caption{\textbf{Comparison to OPDMulti~\cite{sun2023opdmulti}.} We perform a qualitative comparison to OPDMulti~\cite{sun2023opdmulti} (appeared in 3DV 2024) on the same six images presented in \figref{maskrcnn_supp}. OPDMulti fails in various ways: a) segmentation masks bleed outside of the object, and b) merging of multiple objects into one. Our model produces more accurate segmentation masks (which affect the surface normal, and thus the navigation). } 
\figlabel{opdmulti}
\end{figure*}

\subsubsection{\ul{Sim2Real Behavior Cloning}}
We also compare to a Sim2Real behavior cloning approach, building on top of
prior work~\cite{gupta2023predicting}.  Here, we start with their open-loop
imitation learning method and train it on the articulated objects in
ArtObjSim~\cite{gupta2023predicting}.  The policy consumes an RGB-D image and
outputs whole-body motion plans.  Similar to~\cite{gupta2023predicting}, we
find that in simulation, the learned model outputs plans that coarsely capture
the overall motion but aren't precise enough.  We also deploy predicted plans
onto the Stretch in the real world, but the model struggles further due to the
sim2real gap, achieving a 0\% success rate. 

We try to improve this sim2real policy by fine-tuning it on real world data.
We collect 32 demonstrations using tele-operation for each object used for
developing our pipeline (one drawer and two cabinets), and learn a separate
closed-loop behavior cloning policy for each object type. We initialize the
visual backbone using the open-loop policy trained in simulation on ArtObjSim.
We find that these policies only achieve a 70\% success rate on the
\textit{training objects} in the real world, and completely fails on unseen
objects.
A diffusion policy~\cite{chi2023diffusionpolicy} trained on this data also
similarly struggled to generalize to novel objects in previously unseen
environments.

\subsection{Evaluation of Individual Modules}
\seclabel{individual_module_eval}

\subsubsection{\ul{Articulation Parameter Prediction Accuracy}}
\seclabel{apm-eval}
We evaluate 
\model on all of the real world objects in our test
set. We compute a) the detection accuracy, b) handle orientation accuracy, c)
articulation type accuracy,  d) mean handle location error, and e) mean radius
error. The latter four are computed only for detected instances.  We manually
annotate to obtain 3D ground truth for
evaluation.

We find that detection accuracy is high, but not 100\%. 
\model fails to detect 2 of the 31 instances, see example missed instances in \figref{failures}. 
Articulation type prediction is always correct and the handle orientation is
correct 28/29 times.
Furthermore, the mean handle location error is small at $2.11cm$ across
categories.  The mean radius error is $0.98cm$ across categories, which
contrary to our original belief, did not prove to be an obstacle during
deployment. 
\tableref{maskrcnn_table} breaks down results by categories and 
the video
provides qualitative visualizations.
The model detects fully visible objects well, and is sometimes also
able to detect occluded objects.

In addition to our primary evaluations, we also compare \model to two recent works: AO-Grasp \cite{morlans2023aograsp}
and OPDMulti \cite{sun2023opdmulti}.  
We perform a direct quantitative comparison to AO-Grasp to assess differences in handle prediction accuracy.
\tableref{ao-grasp-eval} reports a quantitative comparison of the mean handle
error on our real world images (lower is better). Our method significantly
outperforms AO-Grasp. See \secref{rebuttal_comparisons} for a qualitative
comparison to AO-Grasp, and for more details. 
We also perform a qualitative comparison to OPDMulti (see \figref{opdmulti}),
and find that our model produces more accurate segmentation masks (which affect
the surface normal, and thus the navigation). OPDMulti does not predict handle
locations, disqualifying it as a replacement to APM, and preventing us from
performing a quantitative comparison.
Additionally, we also compare to an approach which directly predicts articulation parameters in 3D (\secref{three_d_comparison}), as well as to Grounded SAM 2 (\secref{grounded_sam_2}).

Since our system is modular, and because we only require 2D predictions for handles and extent, we are able to replace the modified Mask RCNN with another perception model.
In particular, we replace it with Detic, a foundation model for open-vocabulary detection \cite{zhou2022detecting}.
While Detic can produce a segmentation mask for a given object and its handle, it doesn't natively output the handle orientation. 
We design a simple decision rule based on the X and Y variance of the 3D points in the handle segment (in the base robot frame) to obtain the handle orientation. 
We find our Detic-based pipeline to work comparably to our Mask RCNN-based pipeline (see \tableref{maskrcnn_table}).

\begin{table}
  \centering
  \resizebox{\linewidth}{!}{
  \begin{tabular}{lrrrr}
  \toprule
                              & Drawer & Left-Hinged & Right-Hinged & Total\\
  \midrule
  w/ contact correction  & 8/9    & 4/9         & 7/13         & 19/31 \\
  w/o contact correction & 1/9    & 4/9         & 6/13         & 11/31 \\
  \bottomrule
  \end{tabular}}
  \caption{\textbf{Ablations.} Success rates for our full pipeline, both with and without contact correction, evaluated across all objects in the real world. Incorporating contact correction significantly improves the success rate.}
  \tablelabel{ablation}
\end{table}

\subsubsection{\ul{Effectiveness of Adaptation Strategies}}
\seclabel{adaptation-ablation}
\textit{Contact Correction:} As noted in \tableref{ablation}
inclusion of the contact correction adaptation strategy improves success rate
from 35\% to 61\%. 
\figref{contact} visualizes how contact correction aids grasping.
All of this improvement comes from horizontal handles
(primarily on drawers). 
We find contact correction to be a simple but effective strategy 
to tackle last-centimeter errors in grasping.

\begin{figure}[t]
\insertW{0.5}{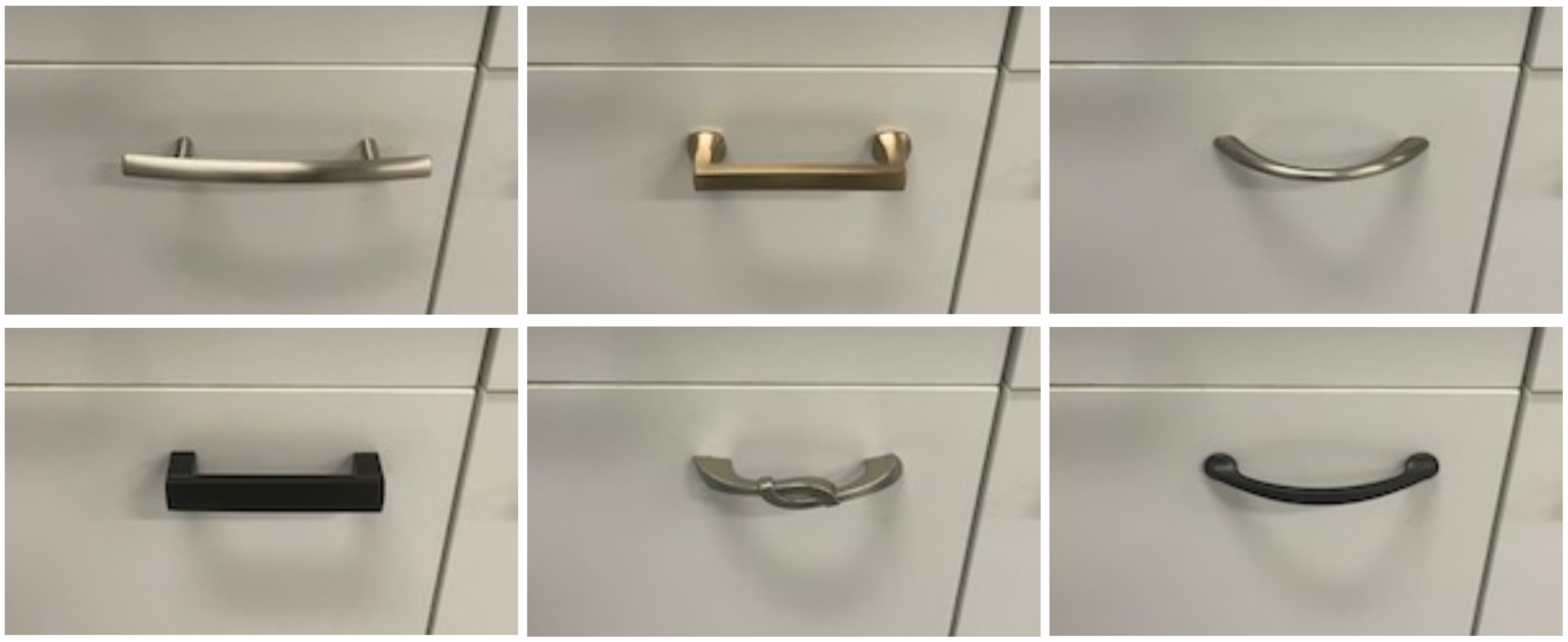}
\caption{\textbf{Diverse Handles.} We test \system on 6 diverse handles on 3  
test objects. \system succeeds on all 18 trials.}
\figlabel{handles_fig}
\end{figure}

\subsection{Evaluation of Interaction between Perception and Control}
\seclabel{robustness}
While the radius estimate from \model was off by $1cm$ on average,
\system was still able to fully articulate the object in a majority of cases.
A tight grasp provides a certain level of tolerance to inaccuracies in state
estimates. Here, we systematically test this.
We perturb the ground truth radius in increments of $2cm$, compute a motion
plan with this inaccurate radius, and deploy this inaccurate plan.
We use ground truth parameters for everything except the radius
estimate to ensure a tight grasp of the handle. 
To our surprise, even with a $10cm$ error in the radius estimate, the robot is able 
to open the cabinet significantly. When the radius is overestimated by $10cm$, the 
cabinet opens to $75^\circ$, and even when underestimated by the same amount, it 
still reaches $55^\circ$, just short of our success threshold.
This suggests that with a solid grasp of the handle, execution can succeed even
with relatively inaccurate state estimates. Handle grasping may be more
critical than millimeter-level accuracy in estimating articulation parameters.

\subsection{Generalizing to Other Articulation Types and Handles}
\seclabel{ovens_and_handles}

\subsubsection{\ul{Other Articulation Types}}
\seclabel{ovens}

While \system was developed on cupboards and drawers, we investigate whether
it is general enough to handle other articulation types. Ovens and other
bottom-hinged objects require a downward semi-circular motion to open. %
which is quite distinct from that of cabinets and drawers.
We conduct experiments to study how well our full end-to-end pipeline
(perception, navigation, and execution) fares for opening a toaster oven.

We evaluate our Detic-based pipeline on three \textit{novel} toaster ovens in
five \textit{previously unseen} kitchens, and find that our pipeline achieves a
80\% success rate (our Mask RCNN-based pipeline struggles to detect ovens due
to the relative scarcity in the ArtObjSim dataset).  In the trial that failed,
Detic failed to detect the toaster oven.

\subsubsection{\ul{Diverse Handles}}
\seclabel{handles}

We study whether our full end-to-end system can work on a diverse set of handles, varying in color, geometry, and material.
To evaluate this, we install \textit{six} diverse handles on unseen objects in our test set, as depicted in \figref{handles_fig}.
This is done for three objects in our test set, one of each articulation type (drawer, left-hinged cabinet, right-hinged cabinet), for a total of 18 trials.
We follow the same protocol as our end-to-end testing in \secref{end_to_end}.

Our system is successful on all 18/18 trials. 
Some handles are more challenging than others, particularly ones which curve, due to reduced lateral tolerance during grasping. Nonetheless, we find that our system is able to successfully open unseen objects even with a diverse set of handles.

\subsection{Failure Mode Analysis}
\seclabel{failure_mode_analysis}

While \system is able to solve a majority of
the novel test objects, our large scale evaluation reveals unforeseen failure
in perception, navigation, and execution as summarized in
\figref{failures}. \textit{59\% of failures (\ie 7 failures) are due to perception,}
including various kinds of failures, such as failure to detect meshed cabinets (2/7), 
incorrect handle orientation prediction (1/7), bad radius prediction (2/7), and confusing keyholes for handles (2/7).
These perception errors are due in part to testing on out of distribution
objects. \model is trained on luxury homes from
HM3D~\cite{ramakrishnan2021hm3d}, whereas we mainly test on office buildings
and apartments. \textit{25\% of failures are in execution} wherein
a firm centered grasp on the handle could not be acquired due to slight errors
in robot calibration and navigation. The robot was able to start opening the
object but eventually the handle slipped out of the gripper. 
The remaining
\textit{15\% of failures are due to navigation errors} on floors with thick 
carpets where the robot would audibly strain while rotating. The robot didn't
get into the predicted pre-grasp pose which led to a failed handle grasp.

The modular design of \system and our approach of making 3D predictions via
lifting 2D predictions in \model, allows us to assess if use of foundation
models (\eg Detic~\cite{zhou2022detecting}, an open vocabulary object detector
trained on broad data) can mitigate the errors due to perception (the largest
failure mode). We find that Detic-based \model detects more instances 
than our Mask RCNN-based \model, but fails in other interesting ways.
Thus, building stronger perception models in 
the context of robotics remains a major challenge in deploying such systems to the real world.

\begin{figure}[t]
\insertW{0.5}{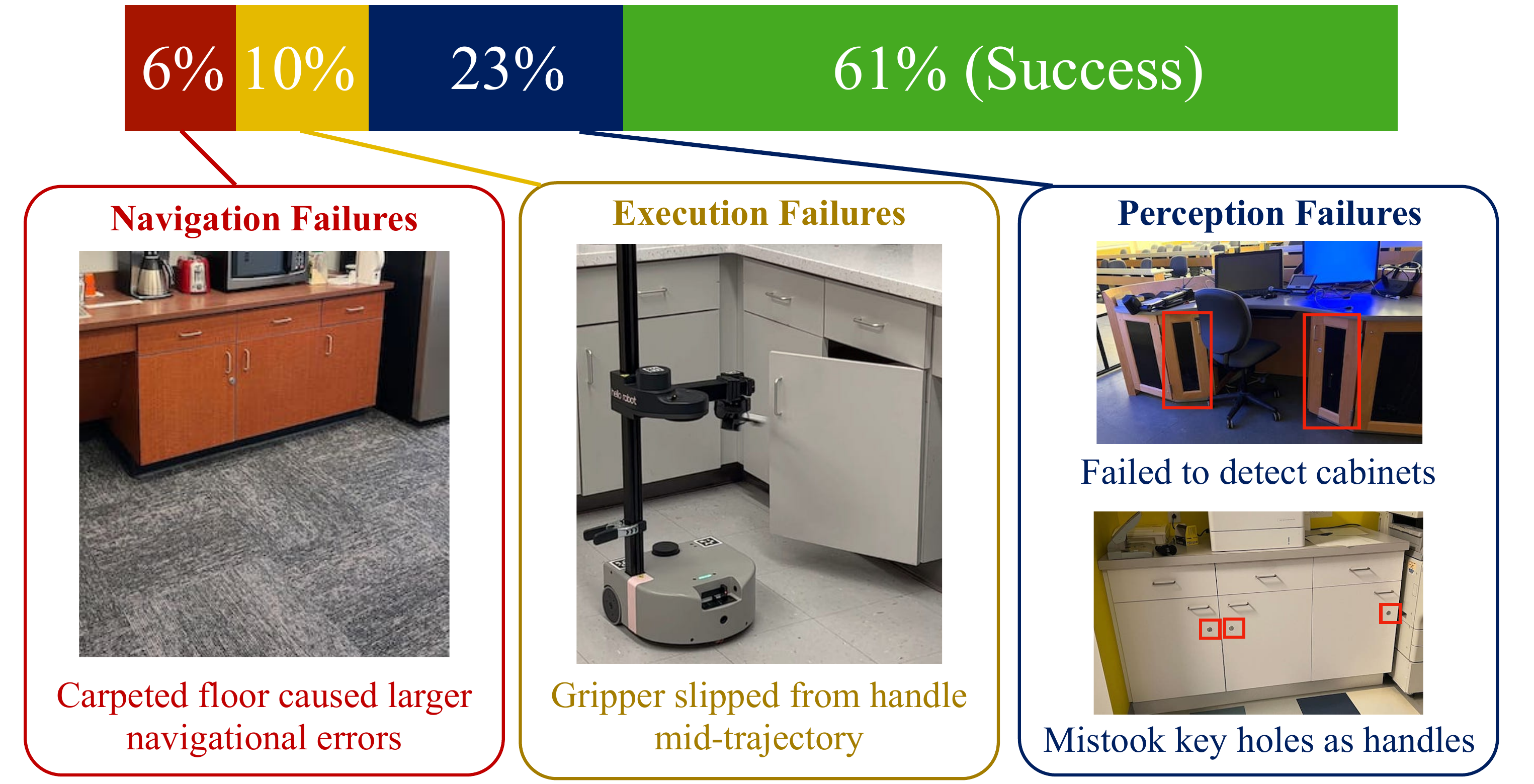}
\caption{\textbf{Failure Cases.} Bar chart characterizing the various failure
modes of \system for opening drawers and cabinets. 59\% of failures are due to
perception failures, including failure to detect the cabinets outlined and
confusing keyholes for handles. Other failures were during execution, where the
handle would slip out, and during navigation, where navigating on carpets was
less accurate than on tiles.}
\figlabel{failures}
\end{figure}

\section{Limitations} 
While we evaluated \system on a variety of objects, our study excluded those with round knobs or no handles. 
\system uses pre-mined navigation targets without allowing for base translation, which may limit performance on objects surrounded by challenging collision geometry or with large-radii.
Additionally, the mapping from articulation parameters to waypoints
must be provided to \system, whereas an end-to-end learned approach may learn this from data.
Finally, there are limitations of the embodiment we use (e.g. it cannot reach cabinets high up, or exert enough force to pull open fridge doors). 

\section{Discussion}
\seclabel{discussion}

\textbullet~Our large-scale experimental study reveals that the modular design
of \system outperforms end-to-end imitation learning methods, even when the 
latter are trained on more than a thousand demonstrations (see
\secref{comparisons}). 
This finding can serve as guidance to practitioners, suggesting that scaling 
imitation-learning datasets beyond 1,000 demonstrations does not necessarily 
lead to generalization across diverse scenarios.
While it is possible that imitation learning
may perform better with even more data, scaling up such effort to collect
even more data remains a challenge. \system's modular design mitigates need for such data
collection.

One might wonder how we contextualize our paper with respect to the Bitter Lesson.
The Bitter Lesson argues that general-purpose methods that scale with data and compute (e.g. end-to-end learning) ultimately outperform systems built with handcrafted structure.
However, our work provides concrete counter evidence: \system is a modular system that outperforms RUM \cite{etukuru2024robot}, a state-of-the-art imitation learning (IL) system by an absolute 35\%. Note, this IL system was already scaled-up: it used 1000+ demos for training. With enough data end-to-end learning \textit{may} prevail, but unlike language and vision domains where data is plentiful, robotics faces unique challenges that raise questions about whether we will \textit{ever} achieve the data volume needed for true generalization.
Simulation could help, but the sim-to-real gap remains a major obstacle.

Furthermore, the bitter lesson may not always hold. For example, the success of NeRFs \cite{mildenhall2020nerf} entirely relies on embedding volumetric rendering into learning—perhaps, modularity is a similar structure to make robotic systems effective.
There is increasing evidence pointing in this direction: \cite{gervet2023navigating} shows that a modular system outperforms a scaled-up IL/RL system for navigation, while \cite{he2024omnih2o} separates perception + high-level reasoning from low-level whole-body control for humanoid control. Furthermore, a modular system doesn’t have to be anti-scaling. Rather, modularity can be very much pro-scaling: each individual module can be scaled up to absorb much more data than a monolithic end-to-end system. For example, the perception module in \system is trained on Internet images, far surpassing the data and diversity that an IL system would be exposed to.
These observations suggest that modularity may not only be compatible with scale, but could be essential for achieving generalization in real world robotics.
  
\textbullet~State-of-the-art perceptions systems for inferring articulation
parameters that have been developed as isolated modules \cite{sun2023opdmulti, 
morlans2023aograsp} do not work well on robot images on in-the-wild articulated objects.
This finding underscores the need for holistic, system-level research to uncover 
previously overlooked bottlenecks that arise during real-world deployment.
We develop \model, which does much better than prior work on previously unseen objects
in novel environments from robot viewpoints in the real world.\\
\textbullet~Modular systems offer distinct advantages over end-to-end learning approaches. 
As vision foundation models advance, a modular system inherently benefits from these improvements without requiring retraining. 
Moreover, we found that adapting our modular system to a new articulation type, such as horizontal-hinged toaster ovens, was straightforward. 
In contrast, an imitation learning system would require collecting additional training data to handle a new articulation type.
These findings suggest that modular architectures not only enhance adaptability to new tasks but also enable systems to leverage ongoing advancements in perception without the overhead of continuous retraining. \\
\textbullet~Perception remains the biggest bottleneck in deploying
such a mobile manipulation system to in-the-wild settings.
The perception failures, even with strong models such as Mask RCNN and Detic, reveal a critical need for broader and more diverse datasets that better capture the variability of real world environments
(\ie everyday cluttered offices / apartments beyond luxury homes). 
This highlights the importance of prioritizing advancements in perception modules to 
accelerate progress toward generalizable mobile manipulation systems.\\
\textbullet~While the system demonstrated robustness to inaccuracies in articulation parameter estimation, successfully opening cabinets even with a 10cm error in radius (\secref{robustness}), achieving an initial secure grasp of the handle remained a critical challenge. 
Grasping failures accounted for approximately 25\% of all observed failures, underscoring the inherent difficulty of achieving precise, last-centimeter adjustments required for successful grasping.
Contact correction provided a partial remedy, but closed-loop visual grasping strategies of some form that can compensate for minor mis-calibrations and imprecise navigation could help further.

These insights offer valuable guidance for researchers and practitioners striving to develop mobile manipulation systems that generalize to unseen objects in real world environments.

\section{Acknowledgments}
This material is based upon work supported by DARPA (Machine Common Sense program), an NSF CAREER Award (IIS-2143873), and the Andrew T. Yang Research and Entrepreneurship Award. We are grateful to the Centre for Autonomy for lending us the Stretch RE2 robot used in this work. We thank Aditya Prakash and Matthew Chang for their feedback on manuscript.

\bibliographystyle{plainnat}

\begin{thebibliography}{90}
\providecommand{\natexlab}[1]{#1}
\providecommand{\url}[1]{\texttt{#1}}
\expandafter\ifx\csname urlstyle\endcsname\relax
  \providecommand{\doi}[1]{doi: #1}\else
  \providecommand{\doi}{doi: \begingroup \urlstyle{rm}\Url}\fi

\bibitem[Abbatematteo et~al.(2020)Abbatematteo, Tellex, and
  Konidaris]{pmlr-v100-abbatematteo20a}
Ben Abbatematteo, Stefanie Tellex, and George Konidaris.
\newblock Learning to generalize kinematic models to novel objects.
\newblock In Leslie~Pack Kaelbling, Danica Kragic, and Komei Sugiura, editors,
  \emph{Proceedings of the Conference on Robot Learning}, volume 100 of
  \emph{Proceedings of Machine Learning Research}, pages 1289--1299. PMLR, 30
  Oct--01 Nov 2020.
\newblock URL \url{https://proceedings.mlr.press/v100/abbatematteo20a.html}.

\bibitem[Ahn et~al.(2022)Ahn, Brohan, Brown, Chebotar, Cortes, David, Finn, Fu,
  Gopalakrishnan, Hausman, Herzog, Ho, Hsu, Ibarz, Ichter, Irpan, Jang, Ruano,
  Jeffrey, Jesmonth, Joshi, Julian, Kalashnikov, Kuang, Lee, Levine, Lu, Luu,
  Parada, Pastor, Quiambao, Rao, Rettinghouse, Reyes, Sermanet, Sievers, Tan,
  Toshev, Vanhoucke, Xia, Xiao, Xu, Xu, Yan, and Zeng]{saycan2022arxiv}
Michael Ahn, Anthony Brohan, Noah Brown, Yevgen Chebotar, Omar Cortes, Byron
  David, Chelsea Finn, Chuyuan Fu, Keerthana Gopalakrishnan, Karol Hausman,
  Alex Herzog, Daniel Ho, Jasmine Hsu, Julian Ibarz, Brian Ichter, Alex Irpan,
  Eric Jang, Rosario~Jauregui Ruano, Kyle Jeffrey, Sally Jesmonth, Nikhil
  Joshi, Ryan Julian, Dmitry Kalashnikov, Yuheng Kuang, Kuang-Huei Lee, Sergey
  Levine, Yao Lu, Linda Luu, Carolina Parada, Peter Pastor, Jornell Quiambao,
  Kanishka Rao, Jarek Rettinghouse, Diego Reyes, Pierre Sermanet, Nicolas
  Sievers, Clayton Tan, Alexander Toshev, Vincent Vanhoucke, Fei Xia, Ted Xiao,
  Peng Xu, Sichun Xu, Mengyuan Yan, and Andy Zeng.
\newblock Do as i can and not as i say: Grounding language in robotic
  affordances.
\newblock In \emph{arXiv preprint arXiv:2204.01691}, 2022.

\bibitem[Bajracharya et~al.(2023)Bajracharya, Borders, Cheng, Helmick, Kaul,
  Kruse, Leichty, Ma, Matl, Michel, Papazov, Petersen, Shankar, and
  Tjersland]{tri_mm_paper}
Max Bajracharya, James Borders, Richard Cheng, Dan Helmick, Lukas Kaul, Dan
  Kruse, John Leichty, Jeremy Ma, Carolyn Matl, Frank Michel, Chavdar Papazov,
  Josh Petersen, Krishna Shankar, and Mark Tjersland.
\newblock Demonstrating mobile manipulation in the wild: A metrics-driven
  approach.
\newblock In \emph{Robotics: Science and Systems XIX}, RSS2023. Robotics:
  Science and Systems Foundation, July 2023.
\newblock \doi{10.15607/rss.2023.xix.055}.
\newblock URL \url{http://dx.doi.org/10.15607/RSS.2023.XIX.055}.

\bibitem[Berenson(2018)]{Berenson2018}
Dmitry Berenson.
\newblock \emph{Obeying Constraints During Motion Planning}, pages 1--32.
\newblock Springer Netherlands, 2018.

\bibitem[Berenson et~al.(2011)Berenson, Srinivasa, and
  Kuffner]{berenson2011task}
Dmitry Berenson, Siddhartha Srinivasa, and James Kuffner.
\newblock Task space regions: A framework for pose-constrained manipulation
  planning.
\newblock \emph{IJRR}, 30\penalty0 (12):\penalty0 1435--1460, 2011.

\bibitem[Burget et~al.(2013)Burget, Hornung, and
  Bennewitz]{Burget2013wbmparticulated}
Felix Burget, Armin Hornung, and Maren Bennewitz.
\newblock {Whole-body motion planning for manipulation of articulated objects}.
\newblock In \emph{ICRA}, pages 1656--1662, 2013.
\newblock ISBN 9781467356411.
\newblock \doi{10.1109/ICRA.2013.6630792}.

\bibitem[Chang et~al.(2020)Chang, Gupta, and Gupta]{chang2020semantic}
Matthew Chang, Arjun Gupta, and Saurabh Gupta.
\newblock Semantic visual navigation by watching youtube videos.
\newblock In \emph{NeurIPS}, 2020.

\bibitem[Chaplot et~al.(2020{\natexlab{a}})Chaplot, Gandhi, Gupta, and
  Salakhutdinov]{chaplot2020object}
Devendra~Singh Chaplot, Dhiraj Gandhi, Abhinav Gupta, and Ruslan Salakhutdinov.
\newblock Object goal navigation using goal-oriented semantic exploration.
\newblock In \emph{In Neural Information Processing Systems (NeurIPS)},
  2020{\natexlab{a}}.

\bibitem[Chaplot et~al.(2020{\natexlab{b}})Chaplot, Gandhi, Gupta, Gupta, and
  Salakhutdinov]{chaplot2020learning}
Devendra~Singh Chaplot, Dhiraj Gandhi, Saurabh Gupta, Abhinav Gupta, and Ruslan
  Salakhutdinov.
\newblock Learning to explore using active neural slam.
\newblock In \emph{International Conference on Learning Representations
  (ICLR)}, 2020{\natexlab{b}}.

\bibitem[Chaplot et~al.(2020{\natexlab{c}})Chaplot, Salakhutdinov, Gupta, and
  Gupta]{chaplot2020neural}
Devendra~Singh Chaplot, Ruslan Salakhutdinov, Abhinav Gupta, and Saurabh Gupta.
\newblock Neural topological slam for visual navigation.
\newblock In \emph{CVPR}, 2020{\natexlab{c}}.

\bibitem[Chen et~al.(2024)Chen, Walsman, Memmel, Mo, Fang, Vemuri, Wu, Fox, and
  Gupta]{chen2024urdformer}
Zoey Chen, Aaron Walsman, Marius Memmel, Kaichun Mo, Alex Fang, Karthikeya
  Vemuri, Alan Wu, Dieter Fox, and Abhishek Gupta.
\newblock {URDFormer}: A pipeline for constructing articulated simulation
  environments from real-world images.
\newblock \emph{arXiv preprint arXiv:2405.11656}, 2024.

\bibitem[Chi et~al.(2023)Chi, Feng, Du, Xu, Cousineau, Burchfiel, and
  Song]{chi2023diffusionpolicy}
Cheng Chi, Siyuan Feng, Yilun Du, Zhenjia Xu, Eric Cousineau, Benjamin
  Burchfiel, and Shuran Song.
\newblock Diffusion policy: Visuomotor policy learning via action diffusion.
\newblock In \emph{Proceedings of Robotics: Science and Systems (RSS)}, 2023.

\bibitem[Chitta et~al.(2010)Chitta, Cohen, and Likhachev]{chitta2010planning}
Sachin Chitta, Benjamin Cohen, and Maxim Likhachev.
\newblock Planning for autonomous door opening with a mobile manipulator.
\newblock In \emph{2010 IEEE International Conference on Robotics and
  Automation}, pages 1799--1806. IEEE, 2010.

\bibitem[Ehsani et~al.(2021)Ehsani, Han, Herrasti, VanderBilt, Weihs, Kolve,
  Kembhavi, and Mottaghi]{ehsani2021manipulathor}
Kiana Ehsani, Winson Han, Alvaro Herrasti, Eli VanderBilt, Luca Weihs, Eric
  Kolve, Aniruddha Kembhavi, and Roozbeh Mottaghi.
\newblock Manipulathor: A framework for visual object manipulation.
\newblock In \emph{CVPR}, pages 4497--4506, 2021.

\bibitem[Eisner* et~al.(2022)Eisner*, Zhang*, and Held]{EisnerZhang2022FLOW}
Ben Eisner*, Harry Zhang*, and David Held.
\newblock Flowbot3d: Learning 3d articulation flow to manipulate articulated
  objects.
\newblock In \emph{Robotics: Science and Systems (RSS)}, 2022.

\bibitem[Etukuru et~al.(2024)Etukuru, Naka, Hu, Lee, Mehu, Edsinger, Paxton,
  Chintala, Pinto, and Shafiullah]{etukuru2024robot}
Haritheja Etukuru, Norihito Naka, Zijin Hu, Seungjae Lee, Julian Mehu, Aaron
  Edsinger, Chris Paxton, Soumith Chintala, Lerrel Pinto, and Nur Muhammad~Mahi
  Shafiullah.
\newblock Robot utility models: General policies for zero-shot deployment in
  new environments.
\newblock \emph{arXiv preprint arXiv:2409.05865}, 2024.

\bibitem[Farshidian et~al.(2017)Farshidian, Jelavic, Satapathy, Giftthaler, and
  Buchli]{Farshidian2017brealtimeplan}
Farbod Farshidian, Edo Jelavic, Asutosh Satapathy, Markus Giftthaler, and Jonas
  Buchli.
\newblock Real-time motion planning of legged robots: A model predictive
  control approach.
\newblock In \emph{ICHR}, pages 577--584, 2017.

\bibitem[Fu et~al.(2022)Fu, Cheng, and Pathak]{fu2022deep}
Zipeng Fu, Xuxin Cheng, and Deepak Pathak.
\newblock Deep whole-body control: Learning a unified policy for manipulation
  and locomotion.
\newblock In \emph{Conference on Robot Learning ({CoRL})}, 2022.

\bibitem[Fu et~al.(2024)Fu, Zhao, and Finn]{fu2024mobile}
Zipeng Fu, Tony~Z Zhao, and Chelsea Finn.
\newblock Mobile aloha: Learning bimanual mobile manipulation with low-cost
  whole-body teleoperation.
\newblock \emph{arXiv preprint arXiv:2401.02117}, 2024.

\bibitem[Gan et~al.(2021)Gan, Schwartz, Alter, Schrimpf, Traer, De~Freitas,
  Kubilius, Bhandwaldar, Haber, Sano, et~al.]{gan2020threedworld}
Chuang Gan, Jeremy Schwartz, Seth Alter, Martin Schrimpf, James Traer, Julian
  De~Freitas, Jonas Kubilius, Abhishek Bhandwaldar, Nick Haber, Megumi Sano,
  et~al.
\newblock Threedworld: A platform for interactive multi-modal physical
  simulation.
\newblock In \emph{Thirty-fifth Conference on Neural Information Processing
  Systems Datasets and Benchmarks Track}, 2021.

\bibitem[Gao et~al.(2023)Gao, Li, Yu, and Shaung]{gao2023twostage}
Fang Gao, XueTao Li, Jun Yu, and Feng Shaung.
\newblock A two-stage fine-tuning strategy for generalizable manipulation skill
  of embodied ai.
\newblock \emph{arXiv preprint arXiv:2307.11343}, 2023.

\bibitem[Gervet et~al.(2023)Gervet, Chintala, Batra, Malik, and
  Chaplot]{gervet2023navigating}
Theophile Gervet, Soumith Chintala, Dhruv Batra, Jitendra Malik, and
  Devendra~Singh Chaplot.
\newblock Navigating to objects in the real world.
\newblock \emph{Science Robotics}, 8\penalty0 (79):\penalty0 eadf6991, 2023.

\bibitem[Gray et~al.(2011)Gray, Clingerman, Likhachev, and Chitta]{GrayPR2}
Steven Gray, Christopher Clingerman, Maxim Likhachev, and Sachin Chitta.
\newblock Pr2: Opening spring-loaded doors.
\newblock In \emph{2010 IEEE/RSJ International Conference on Intelligent Robots
  and Systems}, 2011.

\bibitem[Gray et~al.(2013)Gray, Chitta, Kumar, and Likhachev]{6631117}
Steven Gray, Sachin Chitta, Vijay Kumar, and Maxim Likhachev.
\newblock A single planner for a composite task of approaching, opening and
  navigating through non-spring and spring-loaded doors.
\newblock In \emph{2013 IEEE International Conference on Robotics and
  Automation}, pages 3839--3846, 2013.
\newblock \doi{10.1109/ICRA.2013.6631117}.

\bibitem[Gu et~al.(2023)Gu, Xiang, Li, Ling, Liu, Mu, Tang, Tao, Wei, Yao,
  Yuan, Xie, Huang, Chen, and Su]{gu2023maniskill2}
Jiayuan Gu, Fanbo Xiang, Xuanlin Li, Zhan Ling, Xiqiang Liu, Tongzhou Mu, Yihe
  Tang, Stone Tao, Xinyue Wei, Yunchao Yao, Xiaodi Yuan, Pengwei Xie, Zhiao
  Huang, Rui Chen, and Hao Su.
\newblock Maniskill2: A unified benchmark for generalizable manipulation
  skills.
\newblock In \emph{International Conference on Learning Representations}, 2023.

\bibitem[Gupta et~al.(2023)Gupta, Shepherd, and Gupta]{gupta2023predicting}
Arjun Gupta, Max Shepherd, and Saurabh Gupta.
\newblock Predicting motion plans for articulating everyday objects.
\newblock In \emph{International Conference on Robotics and Automation (ICRA)}.
  IEEE, 2023.

\bibitem[Gupta et~al.(2015)Gupta, Arbel{\'a}ez, Girshick, and
  Malik]{gupta2015aligning}
Saurabh Gupta, Pablo Arbel{\'a}ez, Ross Girshick, and Jitendra Malik.
\newblock Aligning 3d models to rgb-d images of cluttered scenes.
\newblock In \emph{CVPR}, 2015.

\bibitem[He et~al.(2017)He, Gkioxari, Doll{\'a}r, and Girshick]{he2017mask}
Kaiming He, Georgia Gkioxari, Piotr Doll{\'a}r, and Ross Girshick.
\newblock Mask r-cnn.
\newblock In \emph{ICCV}, pages 2961--2969, 2017.

\bibitem[He et~al.(2024)]{he2024omnih2o}
T.~He et~al.
\newblock Omnih2o: Universal and dexterous human-to-humanoid whole-body
  teleoperation and learning.
\newblock In \emph{CoRL}, 2024.

\bibitem[Honerkamp et~al.(2021)Honerkamp, Welschehold, and
  Valada]{honerkamp2021learning}
Daniel Honerkamp, Tim Welschehold, and Abhinav Valada.
\newblock Learning kinematic feasibility for mobile manipulation through deep
  reinforcement learning.
\newblock \emph{IEEE RA-L}, pages 6289--6296, 2021.

\bibitem[Honerkamp et~al.(2023)Honerkamp, Welschehold, and
  Valada]{honerkamp2023learning}
Daniel Honerkamp, Tim Welschehold, and Abhinav Valada.
\newblock N$^2$m$^2$: Learning navigation for arbitrary mobile manipulation
  motions in unseen and dynamic environments.
\newblock \emph{IEEE Transactions on Robotics}, 2023.
\newblock \doi{10.1109/TRO.2023.3284346}.

\bibitem[Huang et~al.(2024)Huang, Wang, Li, Zhang, and Fei-Fei]{huang2024rekep}
Wenlong Huang, Chen Wang, Yunzhu Li, Ruohan Zhang, and Li~Fei-Fei.
\newblock Rekep: Spatio-temporal reasoning of relational keypoint constraints
  for robotic manipulation.
\newblock \emph{arXiv preprint arXiv:2409.01652}, 2024.

\bibitem[Ito et~al.(2022)Ito, Yamamoto, Mori, and Ogata]{ito2022efficient}
Hiroshi Ito, Kenjiro Yamamoto, Hiroki Mori, and Tetsuya Ogata.
\newblock Efficient multitask learning with an embodied predictive model for
  door opening and entry with whole-body control.
\newblock \emph{Science Robotics}, 7\penalty0 (65):\penalty0 eaax8177, 2022.
\newblock \doi{10.1126/scirobotics.aax8177}.
\newblock URL
  \url{https://www.science.org/doi/abs/10.1126/scirobotics.aax8177}.

\bibitem[Jain and Kemp(2009)]{5379532}
Advait Jain and Charles~C. Kemp.
\newblock Pulling open novel doors and drawers with equilibrium point control.
\newblock In \emph{2009 9th IEEE-RAS International Conference on Humanoid
  Robots}, pages 498--505, 2009.
\newblock \doi{10.1109/ICHR.2009.5379532}.

\bibitem[Jain and Kemp(2010)]{jain2010pulling}
Advait Jain and Charles~C Kemp.
\newblock Pulling open doors and drawers: Coordinating an omni-directional base
  and a compliant arm with equilibrium point control.
\newblock In \emph{2010 IEEE International Conference on Robotics and
  Automation}, pages 1807--1814. IEEE, 2010.

\bibitem[Jiang et~al.(2022)Jiang, Mao, Savva, and Chang]{opd}
Hanxiao Jiang, Yongsen Mao, Manolis Savva, and Angel~X. Chang.
\newblock {OPD}: Single-view 3d openable part detection.
\newblock In Shai Avidan, Gabriel Brostow, Moustapha Ciss{\'e}, Giovanni~Maria
  Farinella, and Tal Hassner, editors, \emph{ECCV}, pages 410--426, Cham, 2022.
  Springer Nature Switzerland.
\newblock ISBN 978-3-031-19842-7.

\bibitem[Karayiannidis et~al.(2016)Karayiannidis, Smith, Barrientos, {\"O}gren,
  and Kragic]{karayiannidis2016adaptive}
Yiannis Karayiannidis, Christian Smith, Francisco Eli~Vina Barrientos, Petter
  {\"O}gren, and Danica Kragic.
\newblock An adaptive control approach for opening doors and drawers under
  uncertainties.
\newblock \emph{IEEE Transactions on Robotics}, 32\penalty0 (1):\penalty0
  161--175, 2016.

\bibitem[Kavraki et~al.(1996)Kavraki, Svestka, Latombe, and
  Overmars]{kavraki1996probabilistic}
Lydia~E Kavraki, Petr Svestka, J-C Latombe, and Mark~H Overmars.
\newblock Probabilistic roadmaps for path planning in high-dimensional
  configuration spaces.
\newblock \emph{IEEE transactions on Robotics and Automation}, 12\penalty0
  (4):\penalty0 566--580, 1996.

\bibitem[Kingston et~al.(2018)Kingston, Moll, and
  Kavraki]{kingston2018sampling}
Zachary Kingston, Mark Moll, and Lydia~E Kavraki.
\newblock Sampling-based methods for motion planning with constraints.
\newblock \emph{Annual review of control, robotics, and autonomous systems},
  1:\penalty0 159--185, 2018.

\bibitem[Kolve et~al.(2017)Kolve, Mottaghi, Han, VanderBilt, Weihs, Herrasti,
  Gordon, Zhu, Gupta, and Farhadi]{ai2thor}
Eric Kolve, Roozbeh Mottaghi, Winson Han, Eli VanderBilt, Luca Weihs, Alvaro
  Herrasti, Daniel Gordon, Yuke Zhu, Abhinav Gupta, and Ali Farhadi.
\newblock {AI2-THOR: An Interactive 3D Environment for Visual AI}.
\newblock \emph{arXiv}, 2017.

\bibitem[Kuffner and LaValle(2000)]{kuffner2000rrt}
James~J Kuffner and Steven~M LaValle.
\newblock {RRT-connect}: An efficient approach to single-query path planning.
\newblock In \emph{ICRA}, 2000.

\bibitem[Li et~al.(2019)Li, Wang, Yi, Guibas, Abbott, and
  Song]{li2019articulated-pose}
Xiaolong Li, He~Wang, Li~Yi, Leonidas Guibas, A.~Lynn Abbott, and Shuran Song.
\newblock Category-level articulated object pose estimation.
\newblock \emph{arXiv preprint arXiv:1912.11913}, 2019.

\bibitem[Lin et~al.(2025)Lin, Sachdev, Fan, Malik, and Zhu]{lin2025sim}
Toru Lin, Kartik Sachdev, Linxi Fan, Jitendra Malik, and Yuke Zhu.
\newblock Sim-to-real reinforcement learning for vision-based dexterous
  manipulation on humanoids.
\newblock \emph{arXiv:2502.20396}, 2025.

\bibitem[Liu et~al.(2023)Liu, Mahdavi-Amiri, and Savva]{Liu_2023_ICCV}
Jiayi Liu, Ali Mahdavi-Amiri, and Manolis Savva.
\newblock Paris: Part-level reconstruction and motion analysis for articulated
  objects.
\newblock In \emph{Proceedings of the IEEE/CVF International Conference on
  Computer Vision (ICCV)}, pages 352--363, October 2023.

\bibitem[Liu et~al.(2021)Liu, Lin, Cao, Hu, Wei, Zhang, Lin, and
  Guo]{liu2021swin}
Ze~Liu, Yutong Lin, Yue Cao, Han Hu, Yixuan Wei, Zheng Zhang, Stephen Lin, and
  Baining Guo.
\newblock Swin transformer: Hierarchical vision transformer using shifted
  windows.
\newblock In \emph{ICCV}, pages 10012--10022, 2021.

\bibitem[Lu et~al.(2020)Lu, Merwe, Sundaralingam, and
  Hermans]{lu2020multifingered}
Qingkai Lu, Mark Merwe, Balakumar Sundaralingam, and Tucker Hermans.
\newblock Multifingered grasp planning via inference in deep neural networks:
  Outperforming sampling by learning differentiable models.
\newblock \emph{IEEE Robotics \& Automation Magazine}, PP, 03 2020.
\newblock \doi{10.1109/MRA.2020.2976322}.

\bibitem[Mahler et~al.(2017)Mahler, Liang, Niyaz, Laskey, Doan, Liu, Ojea, and
  Goldberg]{dexnet2}
Jeffrey Mahler, Jacky Liang, Sherdil Niyaz, Michael Laskey, Richard Doan, Xinyu
  Liu, Juan~Aparicio Ojea, and Ken Goldberg.
\newblock Dex-net 2.0: Deep learning to plan robust grasps with synthetic point
  clouds and analytic grasp metrics.
\newblock In \emph{Robotics: Science and Systems (RSS)}, 2017.

\bibitem[McAllister et~al.(2017)McAllister, Gal, Kendall, van~der Wilk, Shah,
  Cipolla, and Weller]{ijcai2017p661}
Rowan McAllister, Yarin Gal, Alex Kendall, Mark van~der Wilk, Amar Shah,
  Roberto Cipolla, and Adrian Weller.
\newblock Concrete problems for autonomous vehicle safety: Advantages of
  bayesian deep learning.
\newblock In \emph{Proceedings of the Twenty-Sixth International Joint
  Conference on Artificial Intelligence, {IJCAI-17}}, pages 4745--4753, 2017.
\newblock \doi{10.24963/ijcai.2017/661}.
\newblock URL \url{https://doi.org/10.24963/ijcai.2017/661}.

\bibitem[Meeussen et~al.(2010)Meeussen, Wise, Glaser, Chitta, McGann, Mihelich,
  Marder-Eppstein, Muja, Eruhimov, Foote, Hsu, Rusu, Marthi, Bradski, Konolige,
  Gerkey, and Berger]{meeussen2010autonomous}
Wim Meeussen, Melonee Wise, Stuart Glaser, Sachin Chitta, Conor McGann, Patrick
  Mihelich, Eitan Marder-Eppstein, Marius Muja, Victor Eruhimov, Tully Foote,
  John Hsu, Radu~Bogdan Rusu, Bhaskara Marthi, Gary Bradski, Kurt Konolige,
  Brian Gerkey, and Eric Berger.
\newblock Autonomous door opening and plugging in with a personal robot.
\newblock In \emph{2010 IEEE International Conference on Robotics and
  Automation}, pages 729--736. IEEE, 2010.

\bibitem[Mildenhall et~al.(2020)Mildenhall, Srinivasan, Tancik, Barron,
  Ramamoorthi, and Ng]{mildenhall2020nerf}
Ben Mildenhall, Pratul~P. Srinivasan, Matthew Tancik, Jonathan~T. Barron, Ravi
  Ramamoorthi, and Ren Ng.
\newblock Nerf: Representing scenes as neural radiance fields for view
  synthesis.
\newblock In \emph{ECCV}, 2020.

\bibitem[Mittal et~al.(2022)Mittal, Hoeller, Farshidian, Hutter, and
  Garg]{mittal2021articulated}
Mayank Mittal, David Hoeller, Farbod Farshidian, Marco Hutter, and Animesh
  Garg.
\newblock Articulated object interaction in unknown scenes with whole-body
  mobile manipulation.
\newblock In \emph{IROS}, 2022.

\bibitem[Mo et~al.(2021)Mo, Guibas, Mukadam, Gupta, and Tulsiani]{Mo_2021_ICCV}
Kaichun Mo, Leonidas~J. Guibas, Mustafa Mukadam, Abhinav Gupta, and Shubham
  Tulsiani.
\newblock Where2act: From pixels to actions for articulated 3d objects.
\newblock In \emph{Proceedings of the IEEE/CVF International Conference on
  Computer Vision (ICCV)}, pages 6813--6823, October 2021.

\bibitem[Morlans et~al.(2023)Morlans, Chen, Weng, Yi, Huang, Heppert, Zhou,
  Guibas, and Bohg]{morlans2023aograsp}
Carlota~Parés Morlans, Claire Chen, Yijia Weng, Michelle Yi, Yuying Huang,
  Nick Heppert, Linqi Zhou, Leonidas Guibas, and Jeannette Bohg.
\newblock Ao-grasp: Articulated object grasp generation.
\newblock 2023.

\bibitem[Morrison et~al.(2018)Morrison, Tow, McTaggart, Smith, Kelly{-}Boxall,
  Wade{-}McCue, Erskine, Grinover, Gurman, Hunn, Lee, Milan, Pham, Rallos,
  Razjigaev, Rowntree, Vijay, Zhuang, Lehnert, Reid, Corke, and
  Leitner]{cartman}
Douglas Morrison, Adam~W. Tow, M.~McTaggart, R.~Smith, N.~Kelly{-}Boxall, Sean
  Wade{-}McCue, J.~Erskine, R.~Grinover, A.~Gurman, T.~Hunn, D.~Lee, Anton
  Milan, Trung Pham, G.~Rallos, A.~Razjigaev, T.~Rowntree, K.~Vijay, Zheyu
  Zhuang, Christopher~F. Lehnert, Ian~D. Reid, Peter Corke, and J{\"{u}}rgen
  Leitner.
\newblock Cartman: The low-cost cartesian manipulator that won the amazon
  robotics challenge.
\newblock In \emph{IEEE International Conference on Robotics and Automation
  (ICRA)}, 2018.

\bibitem[Mousavian et~al.(2019)Mousavian, Eppner, and
  Fox]{mousavian2019graspnet}
Arsalan Mousavian, Clemens Eppner, and Dieter Fox.
\newblock 6-dof graspnet: Variational grasp generation for object manipulation.
\newblock In \emph{International Conference on Computer Vision (ICCV)}, 2019.

\bibitem[Mueller et~al.(2018)Mueller, Dosovitskiy, Ghanem, and
  Koltun]{pmlr-v87-mueller18a}
Matthias Mueller, Alexey Dosovitskiy, Bernard Ghanem, and Vladlen Koltun.
\newblock Driving policy transfer via modularity and abstraction.
\newblock In Aude Billard, Anca Dragan, Jan Peters, and Jun Morimoto, editors,
  \emph{Proceedings of The 2nd Conference on Robot Learning}, volume~87 of
  \emph{Proceedings of Machine Learning Research}, pages 1--15. PMLR, 29--31
  Oct 2018.
\newblock URL \url{https://proceedings.mlr.press/v87/mueller18a.html}.

\bibitem[Narayanan and Likhachev(2015)]{narayanan2015task}
Venkatraman Narayanan and Maxim Likhachev.
\newblock Task-oriented planning for manipulating articulated mechanisms under
  model uncertainty.
\newblock In \emph{2015 IEEE International Conference on Robotics and
  Automation (ICRA)}, pages 3095--3101, 2015.
\newblock \doi{10.1109/ICRA.2015.7139624}.

\bibitem[Nasiriany et~al.(2024)Nasiriany, Maddukuri, Zhang, Parikh, Lo, Joshi,
  Mandlekar, and Zhu]{robocasa2024}
Soroush Nasiriany, Abhiram Maddukuri, Lance Zhang, Adeet Parikh, Aaron Lo,
  Abhishek Joshi, Ajay Mandlekar, and Yuke Zhu.
\newblock {RoboCasa}: Large-scale simulation of everyday tasks for generalist
  robots.
\newblock In \emph{Robotics: Science and Systems (RSS)}, 2024.

\bibitem[Nie et~al.(2022)Nie, Gadre, Ehsani, and Song]{nie2022sfa}
Neil Nie, Samir~Yitzhak Gadre, Kiana Ehsani, and Shuran Song.
\newblock Structure from action: Learning interactions for articulated object
  3d structure discovery.
\newblock \emph{arxiv}, 2022.

\bibitem[Ning et~al.(2023)Ning, Wu, Lu, Mo, and Dong]{ning2023learning}
Chuanruo Ning, Ruihai Wu, Haoran Lu, Kaichun Mo, and Hao Dong.
\newblock Where2explore: Few-shot affordance learning for unseen novel
  categories of articulated objects.
\newblock In \emph{Advances in Neural Information Processing Systems}, 2023.

\bibitem[Pankert and Hutter(2020)]{Pankert2020perceptivempc}
Johannes Pankert and Marco Hutter.
\newblock Perceptive model predictive control for continuous mobile
  manipulation.
\newblock \emph{IEEE RA-L}, pages 6177--6184, 2020.

\bibitem[Peterson et~al.(2000)Peterson, Austin, and Kragic]{peterson2000high}
L~Peterson, David Austin, and Danica Kragic.
\newblock High-level control of a mobile manipulator for door opening.
\newblock In \emph{Proceedings. 2000 IEEE/RSJ International Conference on
  Intelligent Robots and Systems (IROS 2000)(Cat. No. 00CH37113)}, volume~3,
  pages 2333--2338. IEEE, 2000.

\bibitem[Qian and Fouhey(2023)]{qian2023understanding}
Shengyi Qian and David~F Fouhey.
\newblock Understanding 3d object interaction from a single image.
\newblock \emph{arXiv preprint arXiv:2305.09664}, 2023.

\bibitem[Ramakrishnan et~al.(2020)Ramakrishnan, Al-Halah, and
  Grauman]{ramakrishnan2020occant}
Santhosh~K. Ramakrishnan, Ziad Al-Halah, and Kristen Grauman.
\newblock Occupancy anticipation for efficient exploration and navigation,
  2020.

\bibitem[Ramakrishnan et~al.(2022)Ramakrishnan, Chaplot, Al-Halah, Malik, and
  Grauman]{ramakrishnan2022poni}
Santhosh~K. Ramakrishnan, Devendra~Singh Chaplot, Ziad Al-Halah, Jitendra
  Malik, and Kristen Grauman.
\newblock Poni: Potential functions for objectgoal navigation with
  interaction-free learning.
\newblock In \emph{Computer Vision and Pattern Recognition (CVPR), 2022 IEEE
  Conference on}. IEEE, 2022.

\bibitem[Ramakrishnan et~al.(2021)Ramakrishnan, Gokaslan, Wijmans, Maksymets,
  Clegg, Turner, Undersander, Galuba, Westbury, Chang, Savva, Zhao, and
  Batra]{ramakrishnan2021hm3d}
Santhosh~Kumar Ramakrishnan, Aaron Gokaslan, Erik Wijmans, Oleksandr Maksymets,
  Alexander Clegg, John~M Turner, Eric Undersander, Wojciech Galuba, Andrew
  Westbury, Angel~X Chang, Manolis Savva, Yili Zhao, and Dhruv Batra.
\newblock Habitat-matterport 3d dataset ({HM3D}): 1000 large-scale 3d
  environments for embodied {AI}.
\newblock In \emph{Thirty-fifth Conference on Neural Information Processing
  Systems Datasets and Benchmarks Track (Round 2)}, 2021.
\newblock URL \url{https://openreview.net/forum?id=-v4OuqNs5P}.

\bibitem[Ren et~al.(2024)Ren, Liu, Zeng, Lin, Li, Cao, Chen, Huang, Chen, Yan,
  Zeng, Zhang, Li, Yang, Li, Jiang, and Zhang]{ren2024grounded}
Tianhe Ren, Shilong Liu, Ailing Zeng, Jing Lin, Kunchang Li, He~Cao, Jiayu
  Chen, Xinyu Huang, Yukang Chen, Feng Yan, Zhaoyang Zeng, Hao Zhang, Feng Li,
  Jie Yang, Hongyang Li, Qing Jiang, and Lei Zhang.
\newblock Grounded sam: Assembling open-world models for diverse visual tasks,
  2024.

\bibitem[{Ruhr} et~al.(2012){Ruhr}, {Sturm}, {Pangercic}, {Beetz}, and
  {Cremers}]{ruhr2012openingdoors}
T.~{Ruhr}, J.~{Sturm}, D.~{Pangercic}, M.~{Beetz}, and D.~{Cremers}.
\newblock A generalized framework for opening doors and drawers in kitchen
  environments.
\newblock In \emph{ICRA}, pages 3852--3858, 2012.
\newblock \doi{10.1109/ICRA.2012.6224929}.

\bibitem[Scaramuzza et~al.(2014)Scaramuzza, Achtelik, Doitsidis, Friedrich,
  Kosmatopoulos, Martinelli, Achtelik, Chli, Chatzichristofis, Kneip, Gurdan,
  Heng, Lee, Lynen, Pollefeys, Renzaglia, Siegwart, Stumpf, Tanskanen, Troiani,
  Weiss, and Meier]{scaramuzza2014vision}
Davide Scaramuzza, Michael~C. Achtelik, Lefteris Doitsidis, Fraundorfer
  Friedrich, Elias Kosmatopoulos, Agostino Martinelli, Markus~W. Achtelik,
  Margarita Chli, Savvas Chatzichristofis, Laurent Kneip, Daniel Gurdan, Lionel
  Heng, Gim~Hee Lee, Simon Lynen, Marc Pollefeys, Alessandro Renzaglia, Roland
  Siegwart, Jan~Carsten Stumpf, Petri Tanskanen, Chiara Troiani, Stephan Weiss,
  and Lorenz Meier.
\newblock Vision-controlled micro flying robots: From system design to
  autonomous navigation and mapping in gps-denied environments.
\newblock \emph{IEEE Robotics \& Automation Magazine}, 21\penalty0
  (3):\penalty0 26--40, 2014.
\newblock \doi{10.1109/MRA.2014.2322295}.

\bibitem[Schiavi et~al.(2023)Schiavi, Wulkop, Rizzi, Ott, Siegwart, and
  Chung]{schiavi2023learningagentawareaffordancesclosedloop}
Giulio Schiavi, Paula Wulkop, Giuseppe Rizzi, Lionel Ott, Roland Siegwart, and
  Jen~Jen Chung.
\newblock Learning agent-aware affordances for closed-loop interaction with
  articulated objects, 2023.

\bibitem[Schulman et~al.(2014)Schulman, Duan, Ho, Lee, Awwal, Bradlow, Pan,
  Patil, Goldberg, and Abbeel]{schulman2014motion}
John Schulman, Yan Duan, Jonathan Ho, Alex Lee, Ibrahim Awwal, Henry Bradlow,
  Jia Pan, Sachin Patil, Ken Goldberg, and Pieter Abbeel.
\newblock Motion planning with sequential convex optimization and convex
  collision checking.
\newblock \emph{The International Journal of Robotics Research}, 33\penalty0
  (9):\penalty0 1251--1270, 2014.

\bibitem[Shafiullah et~al.(2023)Shafiullah, Rai, Etukuru, Liu, Misra, Chintala,
  and Pinto]{shafiullah2023bringing}
Nur Muhammad~Mahi Shafiullah, Anant Rai, Haritheja Etukuru, Yiqian Liu, Ishan
  Misra, Soumith Chintala, and Lerrel Pinto.
\newblock On bringing robots home.
\newblock \emph{arXiv preprint arXiv:2311.16098}, 2023.

\bibitem[Sleiman et~al.(2021)Sleiman, Farshidian, Minniti, and
  Hutter]{sleiman2021unified}
Jean-Pierre Sleiman, Farbod Farshidian, Maria~Vittoria Minniti, and Marco
  Hutter.
\newblock A unified mpc framework for whole-body dynamic locomotion and
  manipulation.
\newblock \emph{IEEE RA-L}, pages 4688--4695, 2021.

\bibitem[Sleiman et~al.(2023)Sleiman, Farshidian, and Hutter]{Sleiman_2023}
Jean-Pierre Sleiman, Farbod Farshidian, and Marco Hutter.
\newblock Versatile multicontact planning and control for legged
  loco-manipulation.
\newblock \emph{Science Robotics}, 8\penalty0 (81), August 2023.
\newblock ISSN 2470-9476.
\newblock \doi{10.1126/scirobotics.adg5014}.
\newblock URL \url{http://dx.doi.org/10.1126/scirobotics.adg5014}.

\bibitem[Sturm et~al.(2010)Sturm, Jain, Stachniss, Kemp, and Burgard]{5653813}
Jürgen Sturm, Advait Jain, Cyrill Stachniss, Charles~C. Kemp, and Wolfram
  Burgard.
\newblock Operating articulated objects based on experience.
\newblock In \emph{2010 IEEE/RSJ International Conference on Intelligent Robots
  and Systems}, pages 2739--2744, 2010.
\newblock \doi{10.1109/IROS.2010.5653813}.

\bibitem[Sun et~al.(2023)Sun, Jiang, Savva, and Chang]{sun2023opdmulti}
Xiaohao Sun, Hanxiao Jiang, Manolis Savva, and Angel~Xuan Chang.
\newblock {OPDMulti}: Openable part detection for multiple objects.
\newblock \emph{arXiv preprint arXiv:2303.14087}, 2023.

\bibitem[Torne et~al.(2024)Torne, Simeonov, Li, Chan, Chen, Gupta, and
  Agrawal]{torne2024rialto}
Marcel Torne, Anthony Simeonov, Zechu Li, April Chan, Tao Chen, Abhishek Gupta,
  and Pulkit Agrawal.
\newblock Reconciling reality through simulation: A real-to-sim-to-real
  approach for robust manipulation.
\newblock \emph{arXiv preprint arXiv:2403.03949}, 2024.

\bibitem[{Vahrenkamp} et~al.(2013){Vahrenkamp}, {Asfour}, and
  {Dillmann}]{Vahrenkamp2013irm}
N.~{Vahrenkamp}, T.~{Asfour}, and R.~{Dillmann}.
\newblock Robot placement based on reachability inversion.
\newblock In \emph{ICRA}, pages 1970--1975, 2013.
\newblock \doi{10.1109/ICRA.2013.6630839}.

\bibitem[Wang et~al.(2024)Wang, Chen, Yu, Xu, Chen, Fu, Lu, Mu, and
  Luo]{wang2024articulated}
Xi~Wang, Tianxing Chen, Qiaojun Yu, Tianling Xu, Zanxin Chen, Yiting Fu, Cewu
  Lu, Yao Mu, and Ping Luo.
\newblock Articulated object manipulation using online axis estimation with
  sam2-based tracking.
\newblock \emph{arXiv preprint arXiv:2409.16287}, 2024.

\bibitem[Wu et~al.(2022)Wu, Zhao, Mo, Guo, Wang, Wu, Fan, Chen, Guibas, and
  Dong]{wu2022vatmart}
Ruihai Wu, Yan Zhao, Kaichun Mo, Zizheng Guo, Yian Wang, Tianhao Wu, Qingnan
  Fan, Xuelin Chen, Leonidas Guibas, and Hao Dong.
\newblock {VAT}-mart: Learning visual action trajectory proposals for
  manipulating 3d {ART}iculated objects.
\newblock In \emph{International Conference on Learning Representations}, 2022.
\newblock URL \url{https://openreview.net/forum?id=iEx3PiooLy}.

\bibitem[Xie et~al.(2023)Xie, Chen, Chen, Qin, Xiang, Sun, Xu, Wang, and
  Su]{xie2023part}
Pengwei Xie, Rui Chen, Siang Chen, Yuzhe Qin, Fanbo Xiang, Tianyu Sun, Jing Xu,
  Guijin Wang, and Hao Su.
\newblock Part-guided 3d rl for sim2real articulated object manipulation.
\newblock 2023.

\bibitem[Xiong et~al.(2024)Xiong, Mendonca, Shaw, and
  Pathak]{xiong2024adaptive}
Haoyu Xiong, Russell Mendonca, Kenneth Shaw, and Deepak Pathak.
\newblock Adaptive mobile manipulation for articulated objects in the open
  world.
\newblock \emph{arXiv preprint arXiv:2401.14403}, 2024.

\bibitem[Xu et~al.(2022)Xu, Zhanpeng, and Song]{xu2022umpnet}
Zhenjia Xu, He~Zhanpeng, and Shuran Song.
\newblock Umpnet: Universal manipulation policy network for articulated
  objects.
\newblock \emph{IEEE Robotics and Automation Letters}, 2022.

\bibitem[Yang et~al.(2023)Yang, Kim, Kembhavi, Wang, and
  Ehsani]{yang2023harmonic}
Ruihan Yang, Yejin Kim, Aniruddha Kembhavi, Xiaolong Wang, and Kiana Ehsani.
\newblock Harmonic mobile manipulation.
\newblock \emph{arXiv preprint arXiv:2312.06639}, 2023.

\bibitem[Yenamandra et~al.(2023)Yenamandra, Ramachandran, Yadav, Wang, Khanna,
  Gervet, Yang, Jain, Clegg, Turner, et~al.]{yenamandra2023homerobot}
Sriram Yenamandra, Arun Ramachandran, Karmesh Yadav, Austin Wang, Mukul Khanna,
  Theophile Gervet, Tsung-Yen Yang, Vidhi Jain, Alexander~William Clegg, John
  Turner, et~al.
\newblock Homerobot: Open-vocabulary mobile manipulation.
\newblock \emph{arXiv preprint arXiv:2306.11565}, 2023.

\bibitem[Yu et~al.(2024)Yu, Wang, Liu, Hao, Liu, Shao, Wang, and
  Lu]{yu2024gamma}
Qiaojun Yu, Junbo Wang, Wenhai Liu, Ce~Hao, Liu Liu, Lin Shao, Weiming Wang,
  and Cewu Lu.
\newblock Gamma: Generalizable articulation modeling and manipulation for
  articulated objects.
\newblock 2024.

\bibitem[Zeng et~al.(2021)Zeng, Lee, Liang, and
  Kroemer]{zeng2021visualidentificationarticulatedobject}
Vicky Zeng, Tabitha~Edith Lee, Jacky Liang, and Oliver Kroemer.
\newblock Visual identification of articulated object parts, 2021.

\bibitem[Zhang et~al.(2023)Zhang, Eisner, and Held]{zhang2023flowbotplus}
Harry Zhang, Benjamin Eisner, and David Held.
\newblock Flowbot++: Learning generalized articulated objects manipulation via
  articulation projection.
\newblock In \emph{Conference on Robot Learning ({CoRL})}, 2023.

\bibitem[Zhou et~al.(2022)Zhou, Girdhar, Joulin, Kr{\"a}henb{\"u}hl, and
  Misra]{zhou2022detecting}
Xingyi Zhou, Rohit Girdhar, Armand Joulin, Philipp Kr{\"a}henb{\"u}hl, and
  Ishan Misra.
\newblock Detecting twenty-thousand classes using image-level supervision.
\newblock In \emph{ECCV}, 2022.

\bibitem[Zucker et~al.(2013)Zucker, Ratliff, Dragan, Pivtoraiko, Klingensmith,
  Dellin, Bagnell, and Srinivasa]{zucker2013chomp}
Matt Zucker, Nathan Ratliff, Anca~D Dragan, Mihail Pivtoraiko, Matthew
  Klingensmith, Christopher~M Dellin, J~Andrew Bagnell, and Siddhartha~S
  Srinivasa.
\newblock Chomp: Covariant hamiltonian optimization for motion planning.
\newblock \emph{The International Journal of Robotics Research}, 32\penalty0
  (9-10):\penalty0 1164--1193, 2013.

\end{thebibliography}

\clearpage

\renewcommand{\thefigure}{A\arabic{figure}}
\renewcommand{\thetable}{A\arabic{table}}
\renewcommand{\thesection}{A\arabic{section}}

{\Large \bf Appendix}

\section{Supporting Figures and Tables}

\begin{figure}[h]
\insertW{0.5}{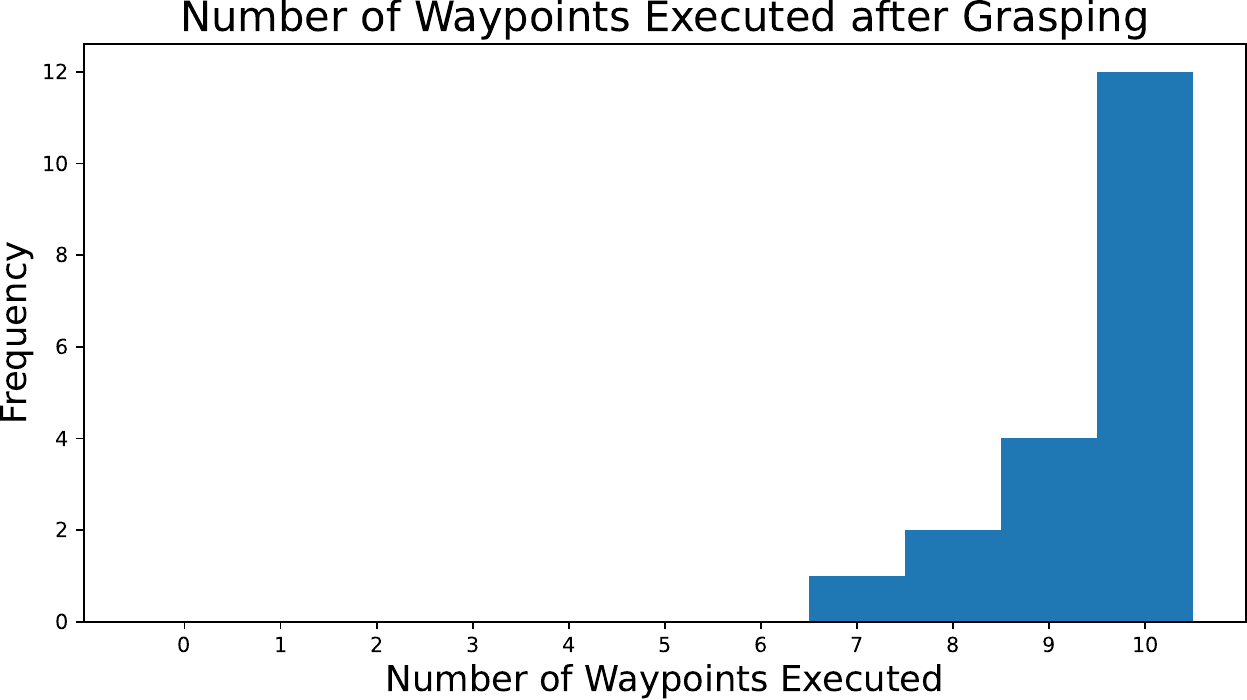}
\caption{\textbf{Number of Waypoints Successfully Executed by \system.} Histogram visualizing the number of waypoints executed, given a grasp of the handle is achieved. For most cases, all 10 waypoints are successfully executed and the object is fully opened.}
\figlabel{waypoints}
\end{figure}

\begin{figure}[h]
\insertW{0.5}{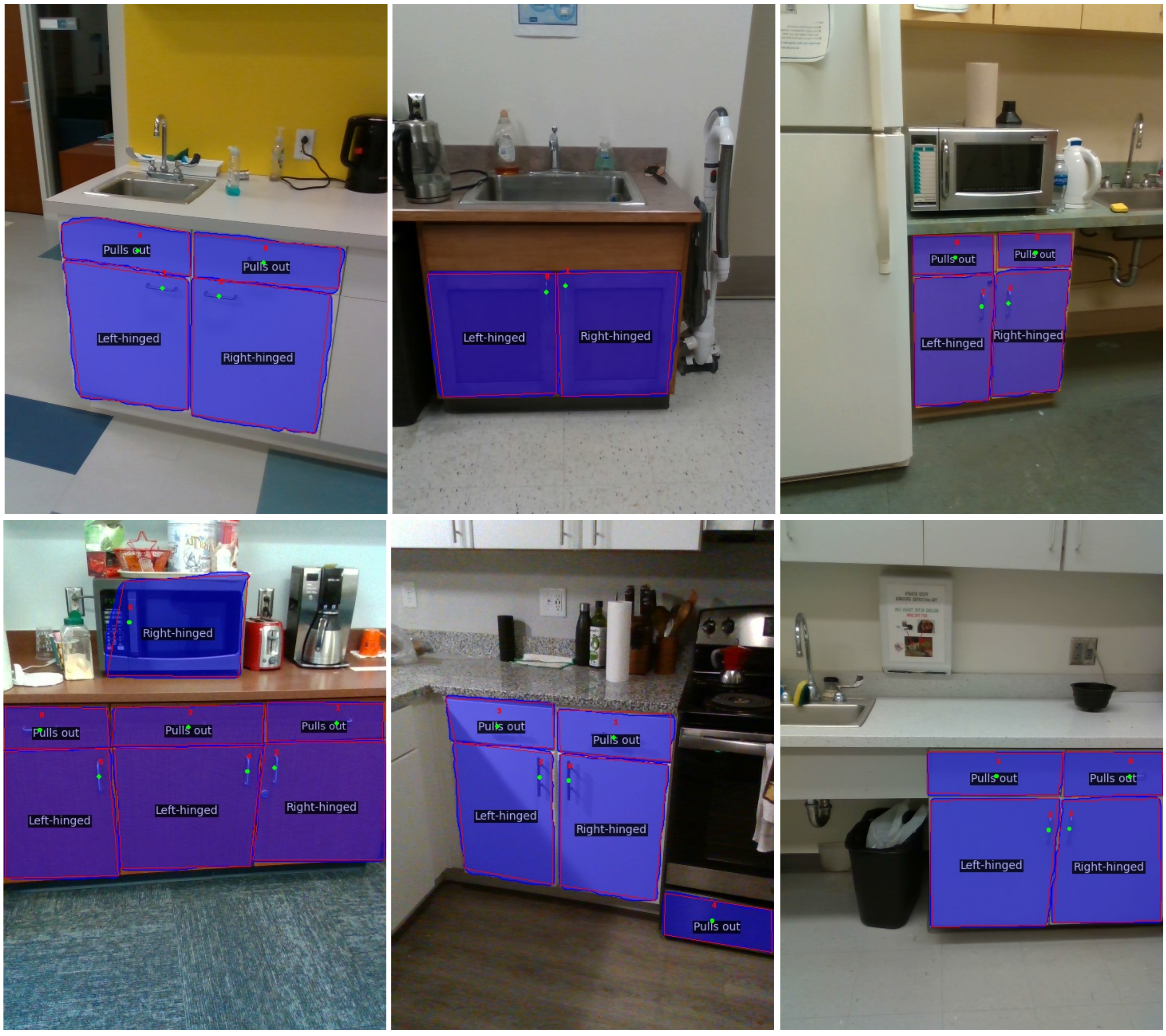}
\caption{\textbf{Qualitative Results from \model.} Output from \model's Mask RCNN on images from our real world testing. We show the predicted segmentation masks in blue, the fitted quadrilaterals in red, the handle point in green, and the predicted articulation type as text on the object.}
\figlabel{maskrcnn_supp}
\end{figure}

\begin{figure*}[t]
\insertW{1.0}{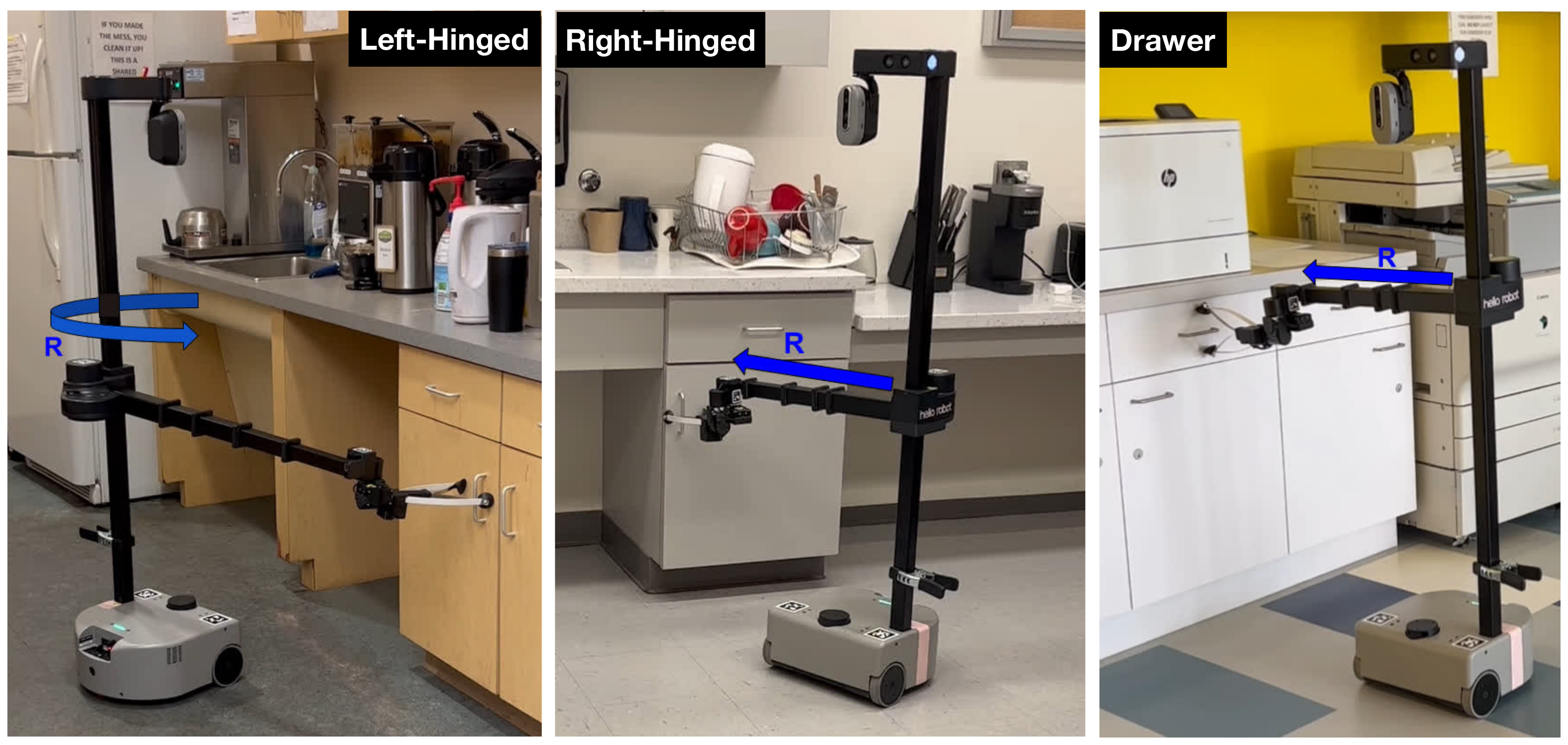}
\caption{{\bf Corrective Motions.} We visualize the corrective motions for the different articulation types. For left-hinged cabinets, this is a counter-clockwise rotation in $1^\circ$ increments. For right-hinged cabinets and drawers, we extend the arm in $1cm$ increments. See \figref{corrective_vertical}(a) and \secref{approach}.}
\figlabel{contact_supp}
\end{figure*}

\clearpage
\newpage

\clearpage
\newpage

\section{Evaluation of Motion Plan Generation}
\seclabel{motion_plan_results}

\begin{figure}[h]
\insertW{0.5}{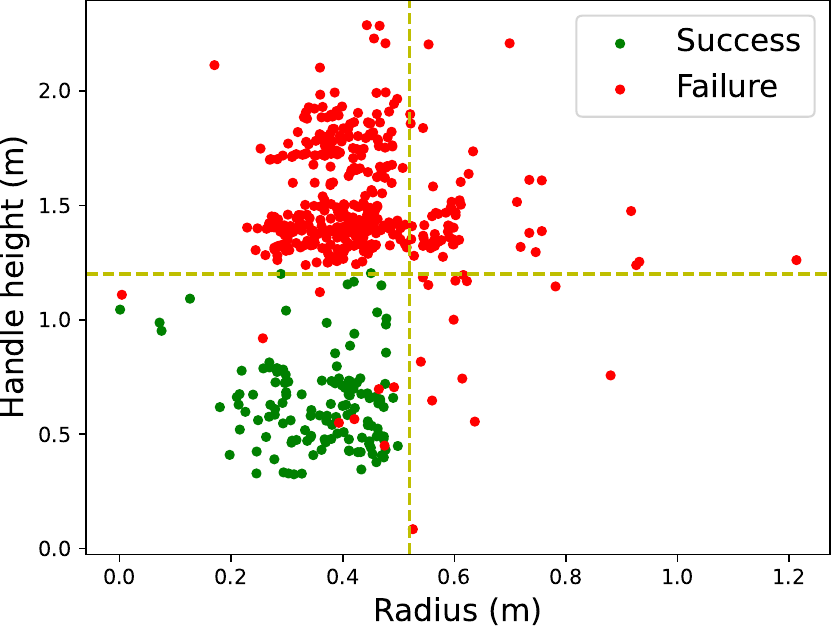}
\caption{\textbf{Effectiveness of Whole-Body SeqIK.} We plot success / failure status of whole-body SeqIK (\secref{approach}) for the cabinets in the ArtObjSim dataset as a function of radius and handle height. We observe that handles above a certain height are difficult due to hardware limitations of the Stretch RE2, and cabinets with a radius larger than a certain threshold are challenging due to our parameterization of motion plans.}
\figlabel{seqik}
\end{figure}

Here, we study the effectiveness of our methodology
for generating whole-body motion plans for opening
objects in the ArtObjSim dataset using the Stretch RE2.
\figref{seqik} plots the success of various cabinets in the 
dataset as a function of both the handle height and the radius.
We notice two trends: cabinets which have handles higher than
$1.2m$ and cabinets with radii larger than $0.5m$ cannot be solved.
The former can be attributed to a hardware limitation of the Stretch RE2.
The lift joint of the robot cannot achieve a height of $1.2m$ or higher, 
leading to this failure.
The latter, on the other hand, can be attributed to our parameterization 
of motion plans. 
Our current formulation only allows for rotation of the base and no translation.
Allowing for the translation of the base in addition to rotation while 
interacting with an object will likely allow the robot to obtain
more complex motion plans, including the kind needed for large-radii cabinets.

\clearpage

\section{Detic~\cite{zhou2022detecting}-based Articulation-parameter Prediction Module}
\seclabel{detic}

Since our system is modular, and because we only require 2D predictions for handles and extent, we are able to replace the modified Mask RCNN with another perception model. We developed a version of \model that uses 2D predictions from Detic~\cite{zhou2022detecting}. Detic is an open-vocabulary object detector trained on broad datasets with an Swin-B backbone~\cite{liu2021swin}.
\tableref{maskrcnn_table} presents perception results with with Detic-based \model.

While Detic can produce a segmentation mask for a given object and its handle, it doesn't natively output the handle orientation. 
We design a simple decision rule based on the X and Y variance of the 3D points in the handle segment (in the base robot frame) to obtain the handle orientation.

Detic detects more instances than Mask RCNN (\tableref{maskrcnn_table}) and is about as accurate in predicting handle location and radius (means are slightly higher due to an outlier prediction but the medians are about the same).

\begin{figure}[h]
\insertW{0.5}{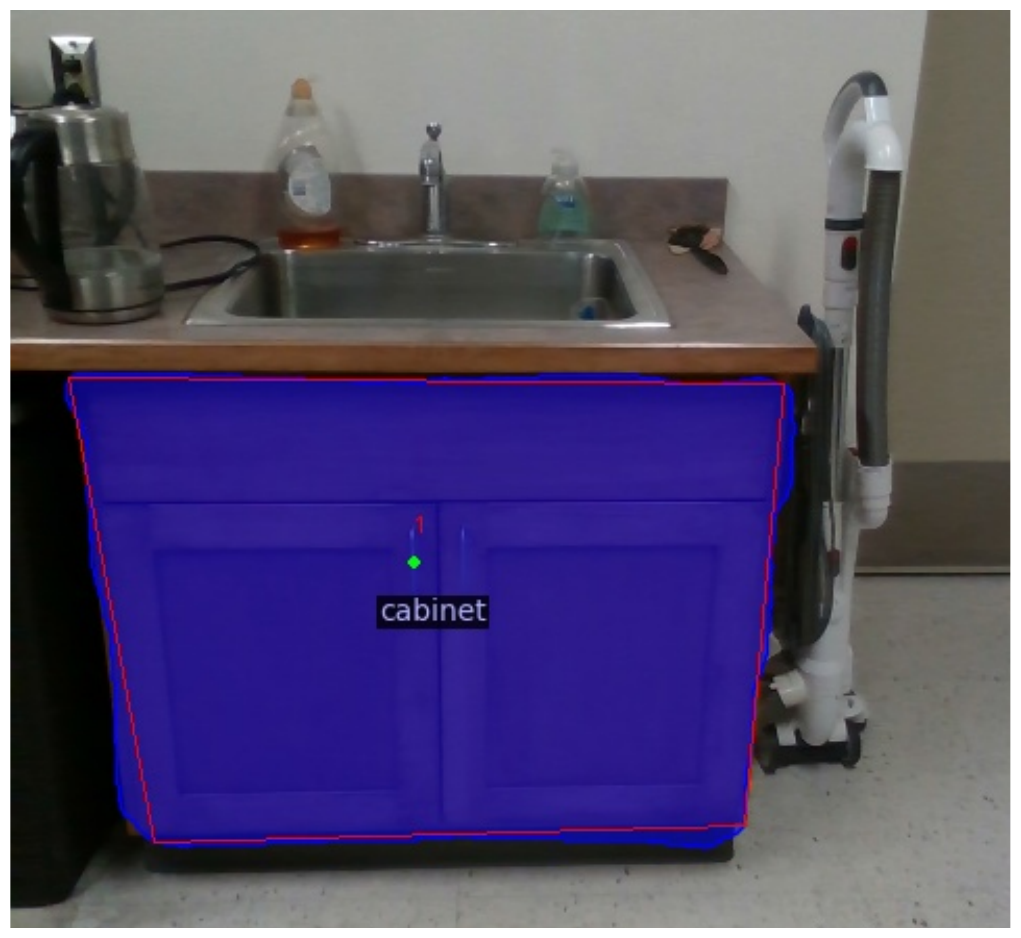}
\caption{\textbf{Detic Failure.} An example failure of using Detic to try and detect articulated objects. In this instance, Detic clumps multiple cabinets together}
\figlabel{detic_failure}
\end{figure}

\clearpage

\section{Other Articulation Types}
\seclabel{ovens}

\begin{figure}[b]
\insertW{0.5}{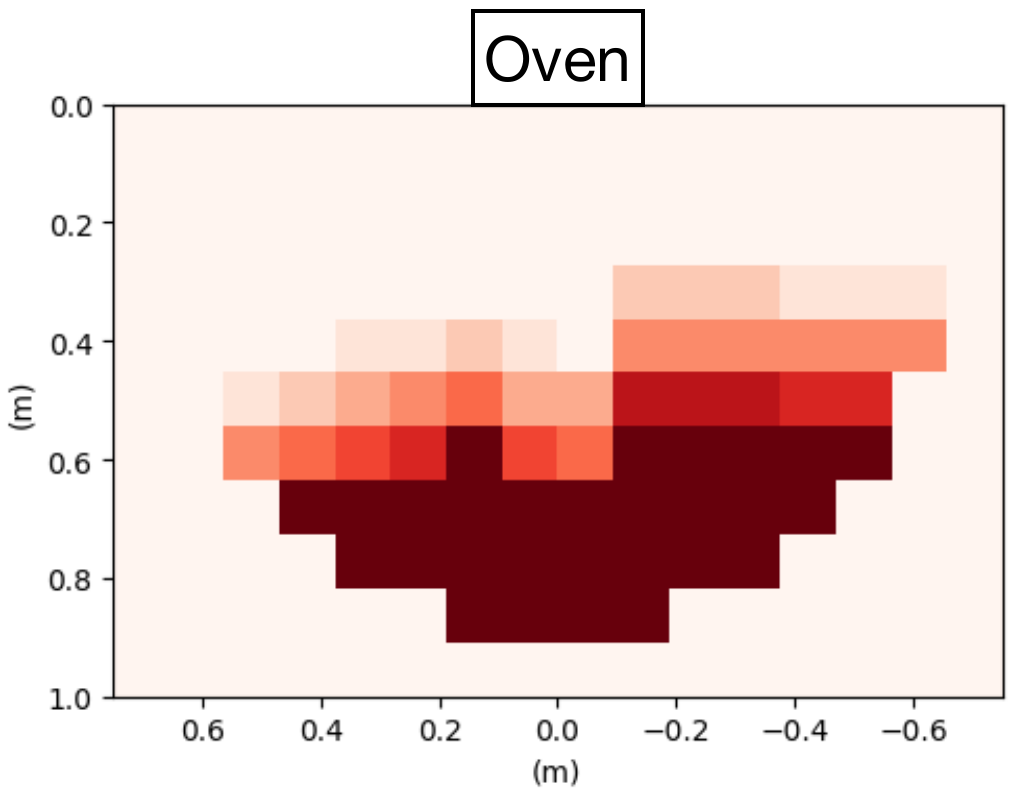}
\caption{\textbf{SeqIK Heatmap for Opening a Bottom-Hinged Object.} Topdown heatmap depicting the success rate of various base locations. The object is placed at (0, 0). Darker red indicates more waypoints were achieved, where all ten waypoints are achieved in the darkest red region.}
\figlabel{oven_heatmap}
\end{figure}

\begin{figure}[h]
\insertW{0.5}{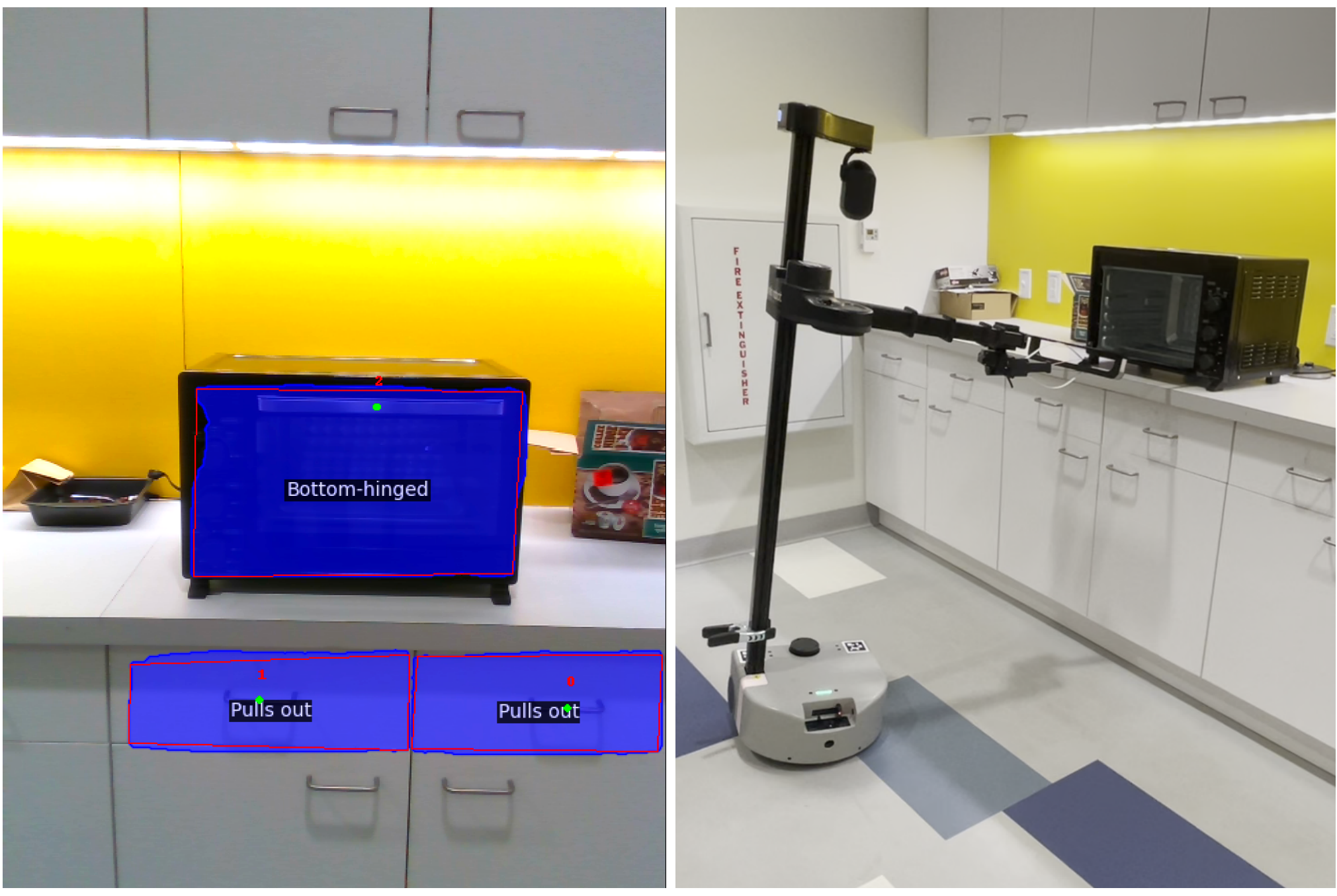}
\caption{\textbf{Perception and Execution for Opening a Toaster Oven.} \textbf{Left.} The output of our modified Mask RCNN demonstrating the detection of the toaster oven. \textbf{Right.} At the end of execution of our full Mask RCNN-based end-to-end pipeline showing the toaster oven is fully open.}
\figlabel{oven_pipeline}
\end{figure}

Even though our focus is opening cupboards and drawers,
we also investigate whether our proposed system is general enough to handle other articulation types. 
Ovens and other bottom-hinged objects require a downward semi-circular motion to open. %
We conduct experiments to check: a) can SeqIK generate motion plans to open bottom-hinged objects with the Stretch, and b)
how well does our full end-to-end pipeline (perception, navigation, and execution) fare for opening a toaster oven?

\noindent \textbf{SeqIK to Generate Motion Plans for Bottom-Hinged Objects.}
We utilize SeqIK to mine motion plans for an object with a bottom-hinge.
As with cabinets and drawers, we limit the motion plans to allow for base rotation and do \textit{not} allow for base translation.
We find that SeqIK is able to find a full motion plan for various different base initializations, as depicted in \figref{oven_heatmap}.

\noindent \textbf{Performance of End-to-End Pipeline.}
We re-train our modified Mask RCNN with the inclusion of ovens and run our entire pipeline, as is, for opening a toaster oven.
\figref{oven_pipeline} shows the output of our modified Mask RCNN model, and the successful opening of a toaster oven using our full end-to-end pipeline.

Across five trials, our modified Mask RCNN-based pipeline achieves a success rate of 20\%. 
In the trials which fail, our modified Mask RCNN either fails to detect the oven, or incorrectly predicts the handle 
to be close to the bottom edge of the detected object.
This is somewhat expected as the ArtObjSim dataset contains only 226 bottom-hinged
objects (ovens, dishwashers, etc.) \vs  1200 drawers and 1992 cabinets. The smaller training set leads to reduced perception performance on such objects.

Since our system is modular, and because we only require 2D predictions for handles and extent, we are able to replace the modified Mask RCNN with another perception model.
In particular, we replace it with Detic, a foundation model for open-vocabulary detection \cite{zhou2022detecting}.
While Detic can produce a segmentation mask for a given object and its handle, it doesn't natively output the handle orientation. 
We design a simple decision rule based on the X and Y variance of the 3D points in the handle segment (in the base robot frame) to obtain the handle orientation. 
Across five trials, our Detic-based pipeline achieves a success rate of 80\%, which is a lot higher than our modified Mask RCNN-based pipeline. This result is somewhat expected: Detic is a much bigger model with a stronger ViT backbone, trained on a much larger dataset than our modified Mask RCNN, which is why it seems to do better at detecting rarer objects.

\clearpage

\newpage
\section{Implementation Details for Baselines}
\seclabel{appendix_comparisons}

\noindent Here, we provide more details on comparisons to end-to-end learning baselines.

\subsection{Robot Utility Models}
\seclabel{rum_details}

We provide additional details about Robot Utility Models (RUM) \cite{etukuru2024robot}.
Our task doesn't provide any privileged information about the object or environment.
RUM makes several assumptions, which we relax in the following ways to enable a comparison in our problem setting:
\begin{itemize}
    \item \textbf{Articulation type.} Firstly, RUM assumes the articulation type as input to the system. For this,
we make use of our articulation type prediction from \model. 
    \item \textbf{Approximate handle height.} Next, RUM requires an approximate height
of the handle to give the eye-in-hand camera an optimal view of the handle. For this as well, we make use of predictions of the 3D handle height from \model. To further benefit RUM, we try out two more heights in the vicinity, and \textit{report the best result for RUM.}
    \item \textbf{Base location.} Finally, RUM assumes the base of the robot is directly positioned in front of the articulated object. To relax this assumption, we allow the robot to navigate to one of two base locations: our pre-mined navigation target, or an `ideal' location which is laterally centered on the handle and 60 cm away from the handle (as our navigation targets, particularly for cabinets, do not necessarily allow for optimal viewpoints of the handle).
    Here as well, we report {\it the best outcome from either navigation targets for RUM}.
\end{itemize}

\subsection{Sim2Real Behavior Cloning}

\begin{figure*}[h]
\insertW{1.0}{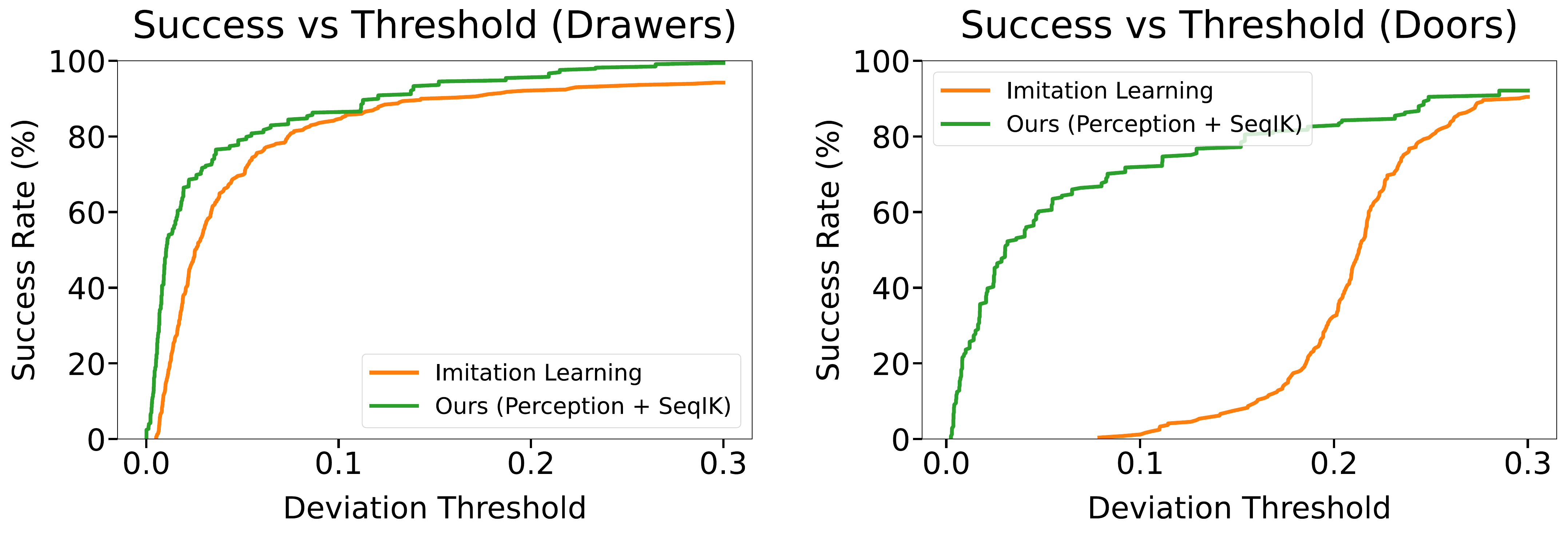}
\caption{\textbf{Simulated Evaluation of Behavior Cloning.} We plot success rate as a function of maximal deviation (meters for translation deviation and radians for rotational deviation). We find that imitation learning particularly struggles to predict accurate robot configurations for cabinet doors.}
\figlabel{sim2real_il}
\end{figure*}

We compare to a pure imitation learning approach from prior work \cite{gupta2023predicting}.
More specifically, we train an open-loop policy which, given an RGB-D image of the object \textit{after} navigation, directly predicts the entire motion plan.
As a result, we train on images simulating the robot base around the target navigation locations for cabinets and drawers in the ArtObjSim dataset.
The camera pose is fixed to look down at the objects (and slightly to the right for the left-hinged cabinets such that they are in view).
As multiple objects may be in view, a red marker is randomly placed somewhere on the object of interest. 
The target motion plan, consisting of $10$ robot configurations corresponding to the $10$ waypoints in the trajectory, is then mined using SeqIK from the target navigation location.

Once the dataset is generated, we learn a policy to predict the motion plan given an RGB-D image by regressing directly to the joint angles.
We experiment with various input modalities (RGB vs RGB-D) and various architectures (fully-connected decoder vs convolutional decoder on top of a pre-trained ResNet encoder) to obtain a policy.
We find that passing in depth and using a convolutional decoder helps achieve the lowest validation loss on held-out objects.

We evaluate the performance of our imitation learning model on unseen objects in simulation.
An image from the unseen validation set is fed through our policy to predict motion plans.
We use forward kinematics to obtain the end-effector pose for each predicted robot configuration, and compute the maximal translational and rotational deviation from the desired end-effector pose. 
For each object, we record the maximal deviation, which could be translational (in meters) or rotational (in radians), with which we plot success rate as a function of deviation threshold in \figref{sim2real_il}.
We compare this to our pipeline in the following manner: we obtain predictions for the articulation parameters using our modified Mask RCNN, and then utilize SeqIK to obtain a motion plan. 
While our method is only slightly better than imitation learning for drawers, our method significantly outperforms imitation learning on cabinet doors.
In particular, the imitation learning model struggles to predict robot configurations with accurate end-effector rotations for the case of cabinet doors.

We evaluate the imitation learning model on a Stretch robot in the real world. 
One object of each articulation type from the test set is tested. 
We find that the model struggles to generate robot configurations which are within joint limits when evaluated on real world data. Even after the joint values are clipped to be within joint limits, the robot is unable to even grasp the handle.
Our hypothesis is that the sim2real gap is exacerbated by noise in the depth images.
Depth images in the real world have random specs and holes close to edges, while the depth images the model is trained on in simulation are a lot cleaner.

\subsection{Real World Behavior Cloning}

\begin{figure}[h]
\insertW{0.5}{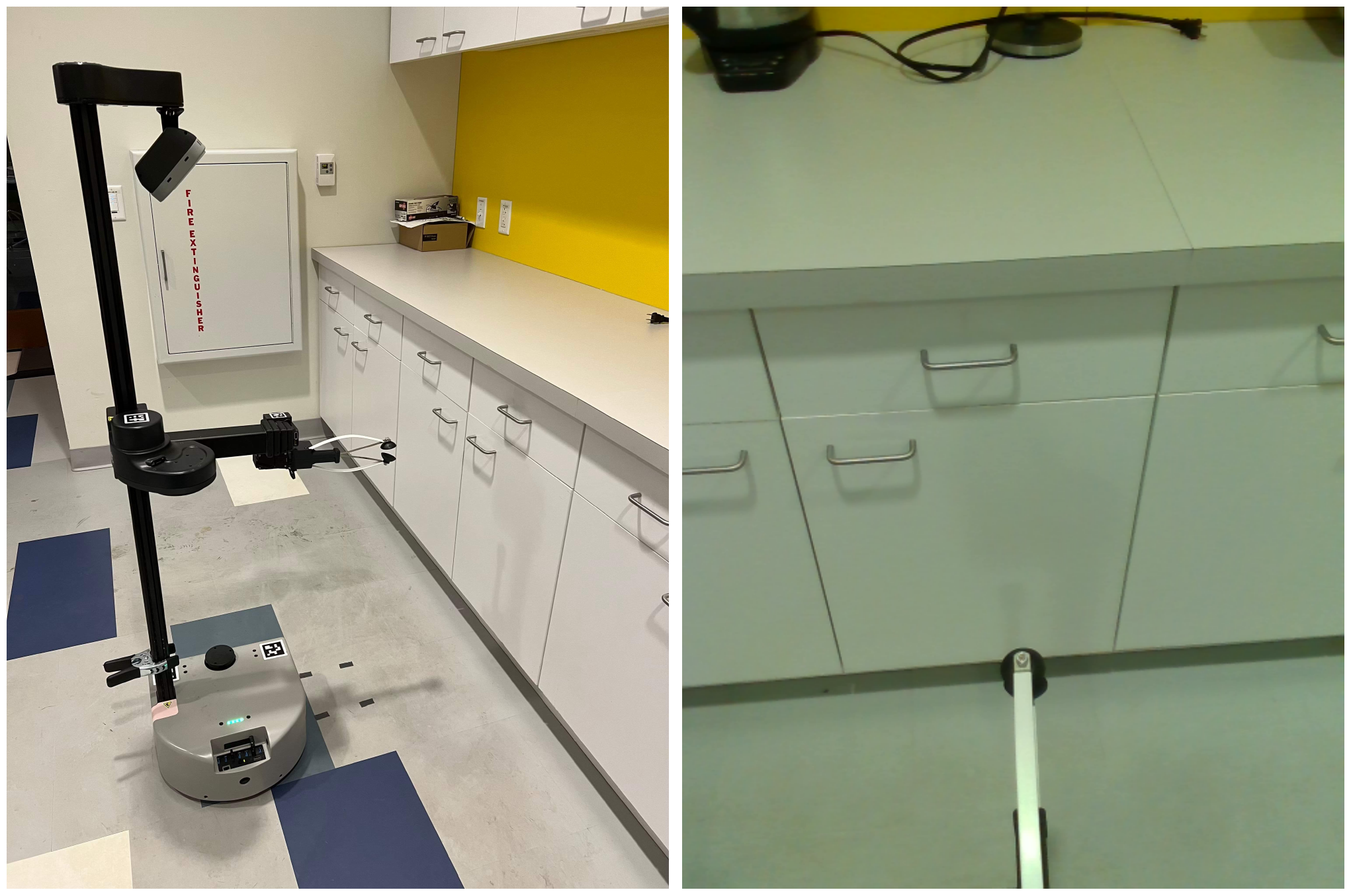}
\caption{\textbf{Real World Behavior Cloning.} \textbf{Left.} The initial configuration of the robot during data collection and testing. \textbf{Right.} The RGB view from the head camera, with both the end-effector and the drawer visible.}
\figlabel{real_world_il}
\end{figure}

With the sim2real failure of our initial imitation learning baseline, we design an imitation learning baseline in the real world. 
We adopt a setup similar to Dobb-E \cite{shafiullah2023bringing}, except we utilize the head camera on the Stretch as opposed to a wrist camera for policy learning.
Here, we collect $32$ demonstrations each for one drawer and one cabinet using tele-operation in the real world. 
We assume access to our pipeline's navigation stack, and therefore collect demonstrations by initializing the robot base in a region surrounding the target navigation location.
This region is represented by a $15\text{cm} \times 15\text{cm}$ grid with $4 \times 4$ base locations, and two demonstrations are collected for each grid point.
To ensure consistency between demonstrations, the robot is set to a pre-determined configuration such that the camera can see both the target handle and the end-effector.
See \figref{real_world_il} for a visualization.

Once the demonstrations have been collected, we learn a closed-loop behavior cloning policy which directly outputs joint angles. 
We experiment with various input modalities (RGB vs RGB-D), various ways of processing the depth image (average pooling vs ResNet encoder), which time-step of action in the future to predict (e.g. 2nd vs 5th action from the current timestep), and whether initializing the policy with the simulation policy helps or not.
The best performing model for the drawer takes in RGB-D, where the depth is processed using average pooling, predicts the 5th action in the future, and is initialized with the simulation policy.
As for the cabinet, the best performing model also utilizes RGB-D with average pooling, and is initialized by the simulation policy, but predicts the 10th action in the future. 
Additionally, we find that filtering out small actions from the dataset helps improve grasping in the real world. This is done by removing samples in each demonstration sequence whose change in joint angles from the previous sample's do not exceed a specified threshold. Input and label pairings are made using these filtered demonstrations.
In summary, both policies leverage ArtObjSim and all real world objects used for development of \system (one drawer and two cabinets), making the comparison to \system fair.

We deploy our behavior cloning policy on a Stretch robot in the real world.
The robot base is initialized to be at an arbitrary location within the $15\text{cm} \times 15\text{cm}$ grid used to collect demonstrations, with the same initial pose and camera orientation.
Across $10$ trials, our best model achieves a $70\%$ success rate at opening the \textit{seen} training drawer used to collect demonstrations. 
The main failure mode is failure to grasp the handle.
We also evaluate our behavior cloning policy on an \textit{unseen} drawer. 
Our best performing model is unable to generalize to this novel drawer, achieving a $0\%$ success rate across $10$ trials.

We repeat the above procedure for a left-hinged cabinet. 
Similar to drawers, our best model is able to open the training cabinet $70\%$ of the time across $10$ trials. 
Here, the failures were caused from failure to grasp the handle and failure to retract the arm even with an accurate grasp pose. 
Furthermore, when evaluated on an \textit{unseen} left-hinged cabinet, the same model achieves a $0\%$ success rate when tested across $10$ trials.

Real world behavior cloning on one object not only fails to generalize to novel objects, but also struggles to consistently open the training object.
This result does not come as a complete surprise given that recent works, which do imitation learning in the real world with a similar number of demonstrations \cite{shafiullah2023bringing}, do not claim generalization, and need to re-train a separate model for each new drawer or cabinet.
It is possible that scaling up imitation learning to a large number of objects may lead to an improvement in performance.
However, real world data collection is not easy to scale up. Collecting $32$ tele-operated demonstrations took roughly 2 hours for the drawer, and even longer for the cabinet due to the increased complexity in the motion plan. This makes large-scale imitation learning prohibitively expensive.

\subsection{Reinforcement Learning} 
Prior work has attempted to use end-to-end reinforcement learning to tackle manipulation tasks such as opening cabinets and drawers.
ManiSkill2 is a recently proposed benchmark and online challenge for generalizable manipulation skills, and includes the task of interacting with drawers and cabinets.
According to the ManiSkill2 Challenge webpage\footnote{https://sapien.ucsd.edu/challenges/maniskill/}, Gao \etal~\cite{gao2023twostage} won the official ManiSkill2 challenge at CVPR 2023's Embodied AI Workshop, placing first on all tracks. 
This method uses end-to-end reinforcement learning with the PPO algorithm for rigid-body tasks like interacting with articulated objects.
Looking at their results (Table III), we see that on the test set, their method achieves a success rate of 24\% on the Open-Cabinet-Door task and 16\% and the Open-Cabinet-Drawer task \textit{in simulation}.
A transfer to the real world of this policy has not been demonstrated, and the sim2real gap seems non-trivial -- the policy is learned on simulated articulated objects in isolation (ie in the absence of a surrounding environment), unlike articulated objects in the real world, making sim2real transfer very difficult.
Furthermore, the success rates of the winning method \textit{in simulation} are significantly lower than the success rates of our method \textit{in the real world}. 

\subsection{Other Prior Systems}
Many other past works use privileged information about the object or environment, or design a method for a custom robot. 
For instance, \cite{5653813} develops a system for opening various household objects including drawers and cabinet doors, but assumes access to privileged information such as the grasping location, the orientation for the end-effector, and the initial pulling direction.
Similarly, \cite{6631117} presents a framework for opening various kinds of doors, but assumes object detection and utilizes a priori knowledge of the object.
\cite{chitta2010planning} develop a graph-search based motion planning algorithm for autonomously opening doors, but also assume access to ground truth articulation parameters.
There are many works of this form which make use of privileged information \cite{jain2010pulling, 5379532, GrayPR2}.
In summary, all of these works make assumptions which simplifies the problem at hand (e.g. assuming access to the kinematic model of the object), do not provide code, or utilize a custom robot, all of which makes comparison to our zero-shot method on a commodity mobile manipulator difficult.

\section{Timing Comparison to RUM}
\seclabel{rum_timing}

\system is significantly faster than RUM \cite{etukuru2024robot}. It executes in 80.68 sec vs. 129.68 sec taken by RUM. \system’s 80.68 secs break down as: perception (7.61 sec), transforms (0.319 sec), navigation (17.33 sec), motion planning (0.091 sec), executing the pre-grasp pose (19 sec), contact correction (12 sec), and the full execution (24.33 sec). Compute is done locally on the Stretch’s Intel i5 processor.

\clearpage
\newpage
\section{Comparisons to Prior Work on Articulation Parameter Estimation}
\seclabel{rebuttal_comparisons}

\begin{figure*}[h!]
\insertW{1.0}{./figures/ao_grasp_figure_v2.pdf}
\caption{\textbf{Comparison to AO-Grasp.} We perform a qualitative comparison to AO-Grasp. AO-Grasp requires a segmented pointcloud – for this, we use APM to segment out objects from our RGB-D images, lift the segmented points to 3D, and then pass this into AO-Grasp to obtain scores for where to grasp. The highest scoring point is treated as the handle location. Our method outperforms AO-Grasp. We believe this is because the handle is not always visible in the noisy point cloud, a drawback of the approach.} 
\figlabel{ao-grasp}
\end{figure*}

Here, we provide more details of our comparison to AO-Grasp \cite{morlans2023aograsp}, as well as a qualitative comparison.
AO-Grasp requires a segmented pointcloud; for this, we use APM to segment out objects from our RGB-D images, lift the segmented points to 3D, and then pass this into AO-Grasp to obtain scores for where to grasp. The highest scoring point is treated as the handle location. 
\figref{ao-grasp} visualizes a qualitative results.
Evidently, the predicted handle is quite off from the ground truth handle in many cases.
We believe this is because the handle may not always visible in the noisy point cloud, a drawback of the approach.

\clearpage
\newpage
\section{Comparison to Direct 3D Prediction}
\seclabel{three_d_comparison}

Additionally, we also compare to an approach which directly predicts articulation parameters in 3D. 
We found that directly predicting the handle in 3D was much worse than our approach (see \tableref{three_d_table}, which reports the mean handle error on our real world images). Our hypothesis for the poor performance of direct 3D prediction is that depth in the real world is quite noisy, making the sim2real transfer of a learned neural network model very difficult.

\begin{table}[h!]
  \centering
  \resizebox{\linewidth}{!}{
  \begin{tabular}{lrrrr}
  \toprule
                              & Drawer & Left-Hinged & Right-Hinged & Total\\
  \midrule
  3D Prediction  & 1.486 m    & 1.711 m  & 1.852 m     &  1.680 m \\
  Ours & 0.0317 m    & 0.0143 m         & 0.0135 m         & 0.0211 m \\
  \bottomrule
  \end{tabular}}
  \caption{\textbf{Quantitative Comparison to direct 3D prediction.} Mean handle error across all instances of our real world testing.}
  \tablelabel{three_d_table}
\end{table}

\section{Comparison to Grounded SAM 2}
\seclabel{grounded_sam_2}

Additionally, we also compare to Grounded SAM 2 \cite{ren2024grounded}, and observe similar trends as Detic (see \tableref{maskrcnn_table}): slightly better mean handle error but higher mean radius error due to occasional clumping of multiple cabinets (e.g. \figref{detic_failure}). Detection, orientation, and articulation accuracies remain high.

\begin{table}[h!]
  \centering
  \resizebox{\linewidth}{!}{
  \begin{tabular}{lrr}
  \toprule
                              & Grounded SAM 2 & Ours \\
  \midrule
  Detection  & 29/31 & 29/31 \\
  Handle orientation & 29/29 & 28/29 \\
  Articulation type & 28/29 & 29/29 \\
  Mean handle error & 1.34 cm & 2.11 cm \\
  Mean radius error & 7.71 cm & 0.98 cm \\
  \bottomrule
  \end{tabular}}
  \caption{\textbf{Comparison to Grounded SAM 2.}}
  \tablelabel{grounded_sam}
\end{table}

\end{document}